\documentclass[journal]{IEEEtran}

\usepackage{cite}
\usepackage{amsmath}
\usepackage{algorithm}
\usepackage{algpseudocode}
\usepackage{array}
\usepackage[caption=false,font=footnotesize]{subfig}
\usepackage{soul}
\usepackage{amsmath}
\usepackage{graphicx}
\usepackage{color}
\usepackage{hyperref}
\usepackage{multirow}
\usepackage{epstopdf}
\usepackage{ulem}
\usepackage{ulem}

\begin{document}
\bstctlcite{bstctl:nodash}

%%%%%%%%% TITLE
\title{A Benchmark for  Edge-Preserving Image Smoothing}

\author{Feida~Zhu,~\IEEEmembership{Student Member,~IEEE,}
        Zhetong~Liang,~\IEEEmembership{Student Member,~IEEE,}
        Xixi~Jia,~\IEEEmembership{Student Member,~IEEE,}
        Lei~Zhang,~\IEEEmembership{Fellow,~IEEE,}
        and~Yizhou~Yu,~\IEEEmembership{Fellow,~IEEE}% <-this % stops a space

\thanks{F. Zhu is with the Department of Computer Science, The University of Hong Kong, Hong Kong.
e-mail: zhufeida@connect.hku.hk.}% <-this % stops a space
\thanks{Z. Liang and L. Zhang are with the Department of Computing, The Hong Kong Polytechnic University, Hong Kong.}
\thanks{X. Jia is with School of Mathematics and Statistics, Xidian University, Xi'an, China.}% <-this % stops a space
\thanks{Y. Yu is with the Department of Computer Science, The University of Hong Kong and Deepwise AI Lab.}
\thanks{This paper has supplementary downloadable material available at http://ieeexplore.ieee.org., provided by the author. The material includes one PDF file of image results. Contact zhufeida@connect.hku.hk for further questions about this work.}
}

%\markright%{IEEE TRANSACTIONS ON IMAGE PROCESSING,~Vol.~14, No.~8, August~2015}%
%{F. Zhu \MakeLowercase{\textit{et al.}}: A Benchmark for  Edge-Preserving Image %Smoothing}

\maketitle
%\thispagestyle{empty}

%%%%%%%%% ABSTRACT
\begin{abstract}
Edge-preserving image smoothing is an important step for many low-level vision problems. Though many algorithms have been proposed, there are several difficulties hindering its further development. First, most existing algorithms cannot perform well on a wide range of image contents using a single parameter setting. Second, the performance evaluation of edge-preserving image smoothing remains subjective, and there lacks a widely accepted datasets to objectively compare the different algorithms. To address these issues and further advance the state of the art, in this work we propose a benchmark for edge-preserving image smoothing. This benchmark includes an image dataset with groundtruth image smoothing results as well as baseline algorithms that can generate competitive edge-preserving smoothing results for a wide range of image contents. The established dataset contains 500 training and testing images with a number of representative visual object categories, while the baseline methods in our benchmark are built upon representative deep convolutional network architectures, on top of which we design novel loss functions well suited for edge-preserving image smoothing. The trained deep networks run faster than most state-of-the-art smoothing algorithms with leading smoothing results both qualitatively and quantitatively. The benchmark will be made publicly accessible.
\end{abstract}

\begin{IEEEkeywords}
Edge-preserving smoothing, Benchmark, Image Dataset, Deep Convolutional Networks
\end{IEEEkeywords}
%%%%%%%%% BODY TEXT

\section{Introduction}
In many image analysis and manipulation tasks, such as contour detection, image segmentation, and image stylization, it is important to preserve major image structures, such as salient edges and contours, while smoothing insignificant details. This can be achieved by edge-preserving image smoothing, a fundamental problem in image processing and low-level computer vision.
%since it is valuable in many image processing and computer graphics tasks, including edge detection, image abstraction, tone manipulation, and detail manipulation~\cite{paris2011local,farbman2008wls,bao2014tree,min2014fast}.
Though a number of algorithms with diverse design philosophies have been proposed \cite{tomasi1998bilateral,perona1990scale,farbman2008edge,fattal2009edge,zhang2014rolling,ham2017robust,xu2011image,min2014fast,bao2014tree,zhang2014100+,paris2011local,bi20151}, there exist three problems that hinder the further development of edge-preserving image smoothing algorithms.

\begin{figure*}[!t]
\captionsetup[subfigure]
  {subrefformat=simple, listofformat=subsimple, labelformat=empty,farskip=1pt}
\centering
\subfloat{\includegraphics[width=0.23\textwidth]{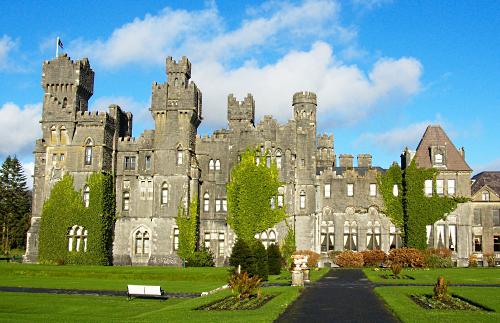}}
\hspace{0.01in}
\subfloat{\includegraphics[width=0.23\textwidth]{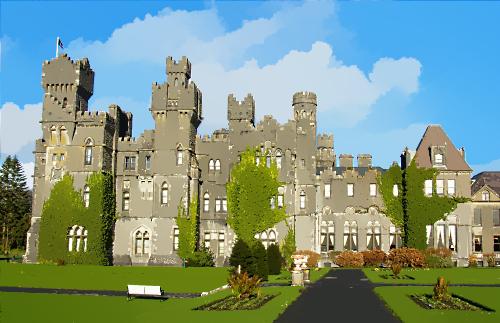}}
\hspace{0.01in}
\subfloat{\includegraphics[width=0.23\textwidth]{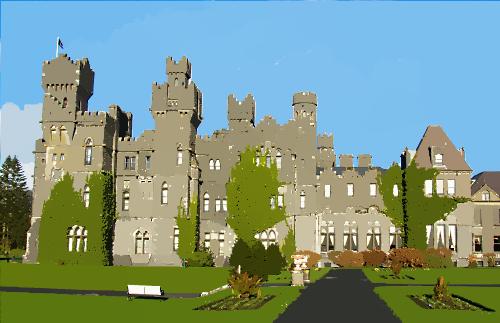}}
\hspace{0.01in}
\subfloat{\includegraphics[width=0.23\textwidth]{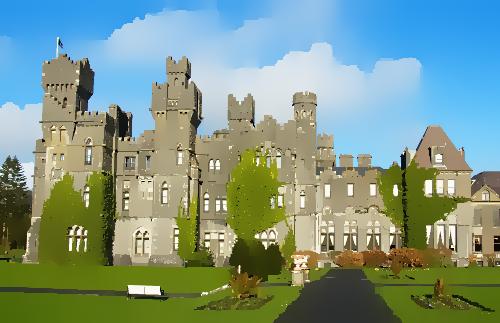}}\
\subfloat{\includegraphics[width=0.23\textwidth]{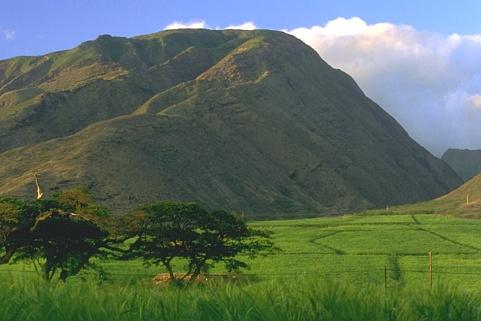}}
\hspace{0.01in}
\subfloat{\includegraphics[width=0.23\textwidth]{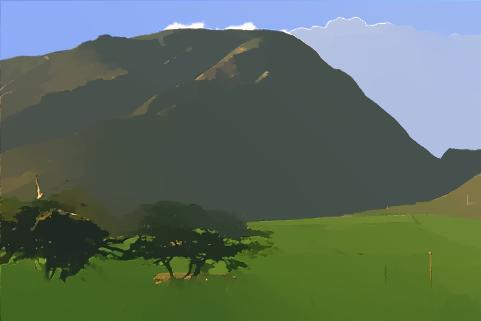}}
\hspace{0.01in}
\subfloat{\includegraphics[width=0.23\textwidth]{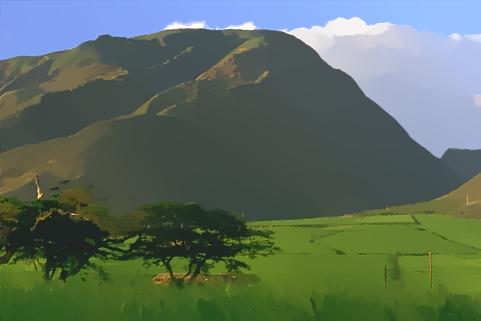}}
\hspace{0.01in}
\subfloat{\includegraphics[width=0.23\textwidth]{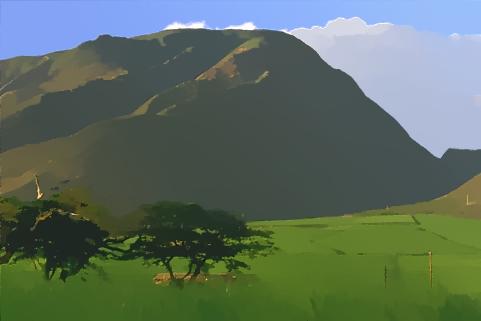}}\
\subfloat{\includegraphics[width=0.23\textwidth]{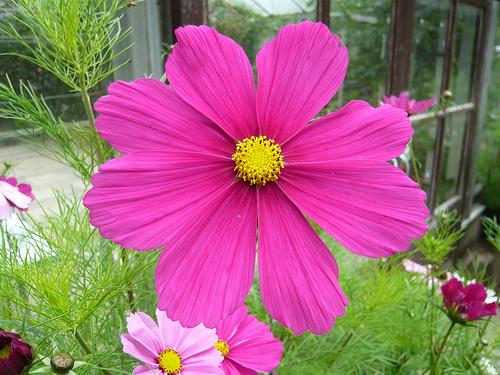}}
\hspace{0.01in}
\subfloat{\includegraphics[width=0.23\textwidth]{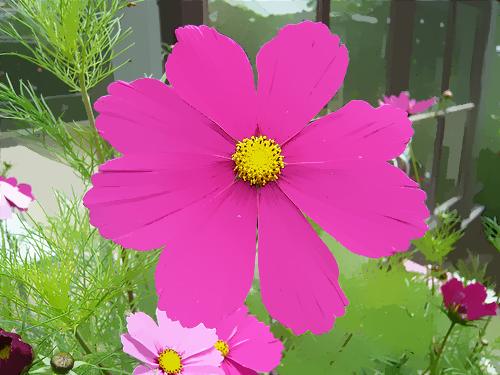}}
\hspace{0.01in}
\subfloat{\includegraphics[width=0.23\textwidth]{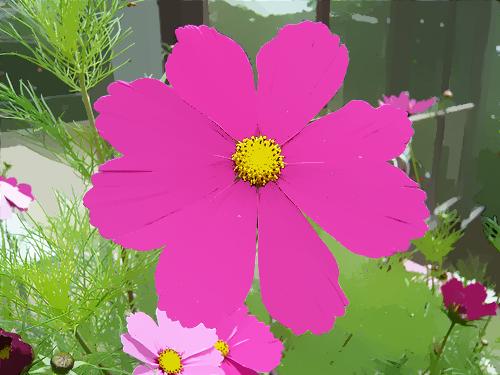}}
\hspace{0.01in}
\subfloat{\includegraphics[width=0.23\textwidth]{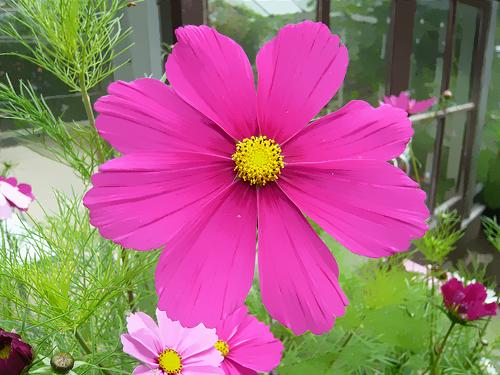}}\
\caption{In our dataset, each source image is associated with 14 edge-preserving smoothing results selected by different subjects from seven edge-preserving smoothing algorithms. The first column shows three source images. On the right three columns, we present 3 human-selected results for each source image.}
\label{figure: introduction figure}
\end{figure*}

First, the performance evaluation of edge-preserving smoothing algorithms remains subjective.  At present, the prevailing method is visual inspection by subjects on the smoothed images. Such an approach is time-consuming and cannot be applied in automatic systems. There lacks an objective metric to evaluate the edge-preserving smoothing algorithms.

Second problem is that an edge-preserving smoothing algorithm is typically evaluated on a very small image set against other algorithms. There lacks a widely accepted large-scale image database for algorithm evaluation. While a smoothing algorithm produces impressive results on certain types of images, it may not perform well on other types of images.  Thus, a large database for a holistic evaluation of edge-preserving smoothing algorithms is much needed.
%Another related issue

Third, smoothing algorithms typically have tunable parameters and images with different categories of contents need different parameter settings. To the best of our knowledge, no smoothing algorithms can perform reasonably well on a wide range of image contents using a single parameter setting.

To address the aforementioned problems, in this paper we propose a benchmark for edge-preserving image smoothing. This benchmark includes an image dataset with ``groundtruth" image smoothing results as well as baseline models that are  capable of generating reasonable edge-preserving smoothing results for a wide range of image contents.
Our image dataset contains 500 training and testing images with a number of visual object categories, including humans, animals, plants, indoor scenes, landscapes and vehicles. The groundtruth smoothing results in our dataset are not directly generated by handcraft approaches, but manually chosen from results generated by existing state-of-the-art edge-preserving smoothing algorithms. This is justified by two reasons. First, as discussed earlier, a single state-of-the-art smoothing algorithm is capable of producing high-quality smoothing results over a small range of image contents especially when its parameters have been fine-tuned. Therefore, a collection of smoothing algorithms are able to generate high-quality results over a wide range of contents. The only caveat is that the best results generated by these algorithms for a specific image need to be hand-picked by humans. Second, since an image has hundreds of thousands of pixels, directly annotating pixelwise smoothing results by humans is too labor-intensive and error-prone.

%To the best of our knowledge, there is temporarily no universally accepted quantitative evaluation method to measure the effectiveness of different smoothing method.  Motivated by the fact that there lack edge-preserved smoothed images ("ground truth") guiding how a source image should be filtered, we construct a dataset with 500 images for training and evaluation. Each natural source image is provided with 14 edge-preserved smoothed result selected by different users from various edge-preserving smoothing algorithms. Figure \ref{figure: introduction figure} shows some examples from the database. To our knowledge, this is the first time that a dataset with human-selected "ground truth" is available for quantitative evaluation.

To establish the baseline algorithms in our benchmark, we resort to the latest deep neural networks. Deep neural networks have a large number of parameters (weights). Once these weights have been trained, they can be fixed and the resulting network has very strong generalization capability and can deal with different types of inputs. Thus, a trained deep neural network on edge-preserving smoothing dataset is expected to perform consistently well in spite of the diverse image contents, which is the goal we want to achieve for edge-preserving image smoothing. We also note that deep learning has been broadly applied to low-level computer vision problems and has achieved state-of-the-art results. Examples include reproducing edge-preserving filters \cite{xu2015deep,liu2016learning,li2016deep}, image denoising \cite{mao2016imagedenoise,zhang2017beyond}, image super-resolution \cite{kim2016accurate,ledig2016photo-realistic,kim2016deeply,Tai-DRRN-2017,Tai-MemNet-2017}, and JPEG deblocking \cite{dong2015compression,zhang2017beyond}. Specifically, we use the following two existing representative network architectures as our baseline methods, very deep convolutional networks (VDCNN) and deep residual networks (ResNet). On top of these network architectures, we design novel loss functions well suited for edge-preserving image smoothing. The deep networks trained over our dataset run faster than most state-of-the-art edge-preserving smoothing algorithms, while the smoothing performance of our ResNet-based model outperforms these algorithms both qualitatively and quantitatively. Our benchmark will be publicly released.

The remainder of this paper is organized as follows. Section 2 reviews the prior work in the research. Section 3 describes the construction of our dataset in detail. The objective metric for edge-preserving smoothing is presented in Section 4. Section 5 describes our baseline deep learning model for the smoothing task. In Section 6, we verify our benchmark by applying our baseline model to the tone mapping and contrast enhancement tasks. Section 7 is the conclusion.

\section{Related Work}
\noindent \textbf{Edge-Preserving Smoothing:} Many methods have been proposed for on edge-preserving smoothing, which can be categorized into two groups. The first group is local filter based approaches, where the filters are designed based on image statistics within a local window. Representative filters include Bilateral Filter~\cite{tomasi1998bilateral}, Weighted Median Filter (WMF)~\cite{zhang2014100+}, Anisotropic Diffusion (AD)~\cite{perona1990scale}, and Edge-avoiding Wavelet (EAW)~\cite{fattal2009edge}. Rolling Guidance Filter (RGF)~\cite{zhang2014rolling} applies the weighted filters iteratively, and Tree Filtering~\cite{bao2014tree} utilizes minimum spanning tree to smooth out details while preserving major structure. However, the local filters have a common limitation in that they often introduce artifacts (such as halos along the edge) because only the local image statistics are used in the filtering, and we cannot explicitly control the statistical properties of the filtered images.

The second group is global optimization based approaches. The smoothed image is obtained by solving a global objective function, which usually involves a data term, constraining the distance between original image and smoothed image, and a regularization term, striving to achieve smoothness. Representative methods include Weighted Least Square smoothing (WLS)~\cite{farbman2008edge}, $L_0$ smoothing~\cite{xu2011image}, Fast Global Smoother (FGS)~\cite{min2014fast}, $L_1$ smoothing~\cite{bi20151}  and SD filter~\cite{ham2017robust}. Such methods overcome several limitations of local filter based approaches such as halos and gradient reversals. However, increased computational cost comes with solving the large-scale linear systems\cite{farbman2008edge,bi20151}.

%These algorithms generally aim to avoid edge blurring, halo artifacts, gradient reversal, and global intensity shift.

\noindent \textbf{Quantitative Evaluation:} Bao \textit{et al}.~\cite{bao2014tree} applied different edge-preserving filters to the test images in Berkeley Segmentation Dataset (BSDS300)~\cite{MartinFTM01} prior to boundary detection. F-measure is used to evaluate the filters' effectiveness in suppressing trivial details and preserving edges. Ham \textit{et al}.~\cite{ham2015robust,ham2017robust} proposed ODS and OIS~\cite{arbelaez2011contour} using the gradient magnitudes of filtered images to measure the effectiveness of filters. However, these evaluation approaches may suffer from the deviation of detected boundaries. In contrast, we construct a dataset of source images and their associated ``groundtruth" (human-selected edge-preserving smoothed images). Objective evaluation can be performed by directly comparing the results of an edge-preserving filter against such ``groundtruth".

\noindent \textbf{Deep Edge-aware Filter Learning:} Xu \textit{et al}.~\cite{xu2015deep} proposed a method to learn and reproduce individual edge-preserving filters. They used a convolutional neural network to predict smoothed image gradients and then run an expensive step to reconstruct the smoothed image itself. However, the $\beta$ parameter in their reconstruction step is not fixed and varies with different filters. Liu \textit{et al}.~\cite{liu2016learning} proposed a hybrid network by incorporating spatially varying recurrent neural networks (RNN) conditioned on the input image. A deep CNN is used to learn the weight map of the RNN. Li \textit{et al}.~\cite{li2016deep} proposed a learning-based method to construct a CNN-based joint filter to transfer the structure of a guidance image to a target image for structure-texture separation. Fan \textit{et al}.~\cite{fan2017generic} exploited edge information by separating the image smoothing problem into two steps. The first sub-network is supervised to predict the edge map and the second sub-network reconstructs the target image by leveraging the predicted edge map.

In contrast, we do not aim to reproduce individual filters but the best results among a number of filters over a wide range of image contents. Furthermore, our deep neural networks learn pixelwise colors in the smoothed result instead of smoothed image gradients. Thus an input image can be smoothed efficiently with a single forward pass through the network without the need of a gradient-based reconstruction step.

\section{A Dataset for Edge-Preserving Smoothing}

\begin{figure}[!t]
\centering
\includegraphics[width=0.49\textwidth]{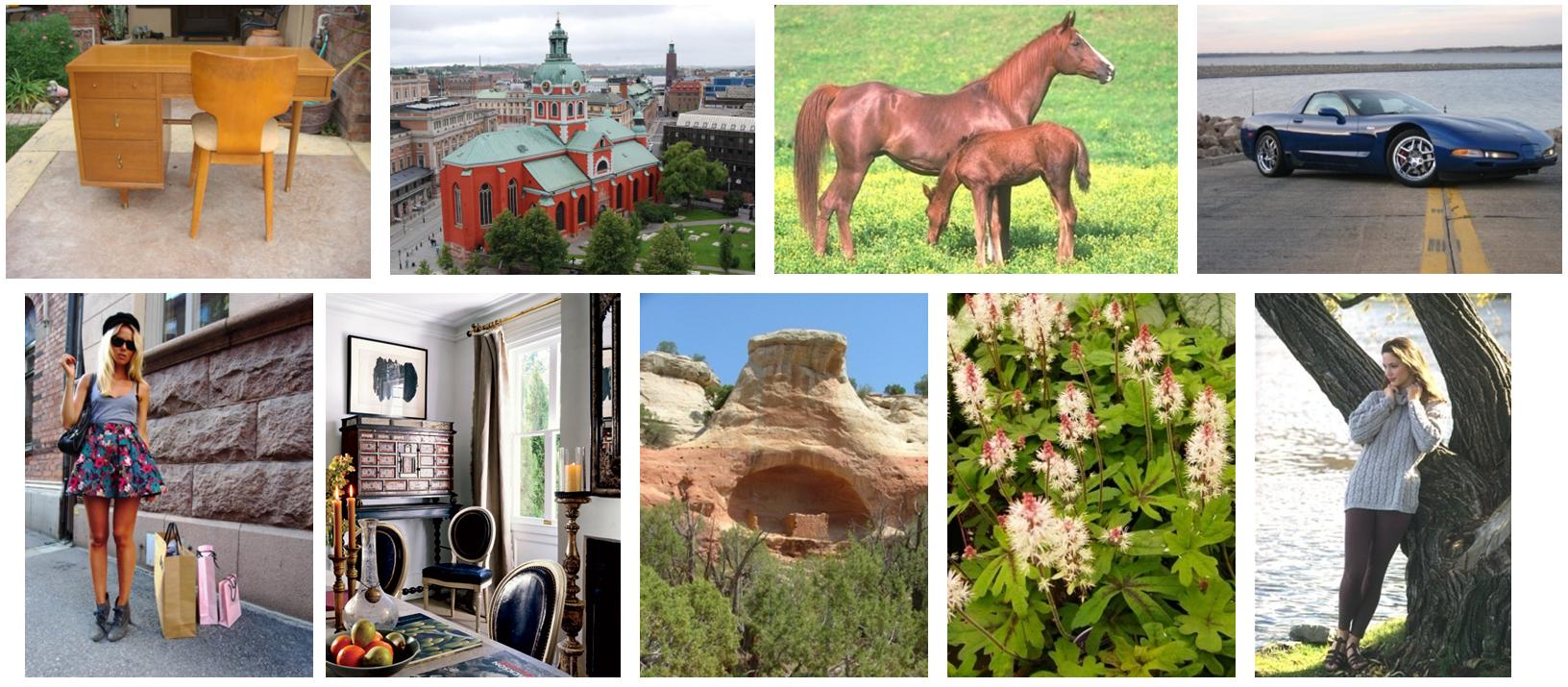}
\caption{Sample source images from our dataset for edge-preserving image smoothing.
}
\label{Fig_dataset_sample}
\end{figure}

A major source of our images  is the database reported in \cite{ma2016group}. This database is composed of a large number of high-quality natural images, which were originally employed for comparing image quality models. Another source of our dataset is the Berkeley Segmentation Dataset (BSDS500)~\cite{arbelaez2011contour} which has been widely used in the computer vision community. We manually chose 500 images of a variety of objects and scenes (humans, animals, plants, indoors, landscape, vehicles, etc.). Some sample images are shown in Figure~\ref{Fig_dataset_sample}.  The selected images in the dataset contain clear structures and visible details which are well suited for evaluating edge preserving smoothing algorithms. Besides, the dataset are balanced among different objects and scenes. The images not suitable for edge-preserving smoothing (e.g., images without clear structures but filled with textured regions) are excluded from our dataset. 

As mentioned earlier, the groundtruth smoothing results in our dataset are not directly annotated by humans, but manually chosen from results generated by existing state-of-the-art edge-preserving smoothing algorithms. This is because an image has hundreds of thousands of pixels, and thus directly annotating pixelwise smoothing results by humans is too labor-intensive and error-prone. On the other hand, while a single state-of-the-art algorithm is capable of producing high-quality smoothing results over a certain range of image contents with fine-tuned parameters, a collection of different smoothing algorithms are able to generate high-quality results over a wide range of contents.

\subsection{Selection Tool}
%The crucial problem is "what should the edge-preserved smoothed image look like". In fact, it's challenging for a user to imagine and create the smoothed result directly from the source image. Choosing from smoothed results of existing methods is an alternative way. Post-processing to the chosen result have not been performed due to limited labor source, which can be implemented in future research.
We chose seven state-of-the-art and representative edge-preserving algorithms to construct our dataset, including SD filter~\cite{ham2017robust}, $L_0$ smoothing~\cite{xu2011image}, Fast Global Smoother (FGS)~\cite{min2014fast}, Tree Filtering~\cite{bao2014tree}, Weighted Median Filter (WMF)~\cite{zhang2014100+}, $L_1$ smoothing~\cite{bi20151} and Local Laplacian filter (LLF)~\cite{paris2011local}. The selection considerations of these filters are twofold. The first is the representativeness and impact of the work (e.g., high citations). This consideration ensures that the selected filters are state-of-the-arts. The second consideration is the diversity to leverage the merits of different types of filters. Based on these considerations, we selected 4 global methods \cite{ham2017robust,xu2011image,min2014fast,bi20151} and 3 local methods \cite{bao2014tree,zhang2014100+,paris2011local}. The global filters explicitly formulate the edge-preserving smoothing process as a global optimization problem, while the local filters apply a weighted function depending on the similarity of features within a local window.

\begin{figure}[!t]
%\captionsetup[subfigure]{subrefformat=simple, listofformat=subsimple, labelformat=empty}
\centering
\subfloat[Step 1: Choose the best result for each algorithm]{\includegraphics[width=0.48\textwidth]{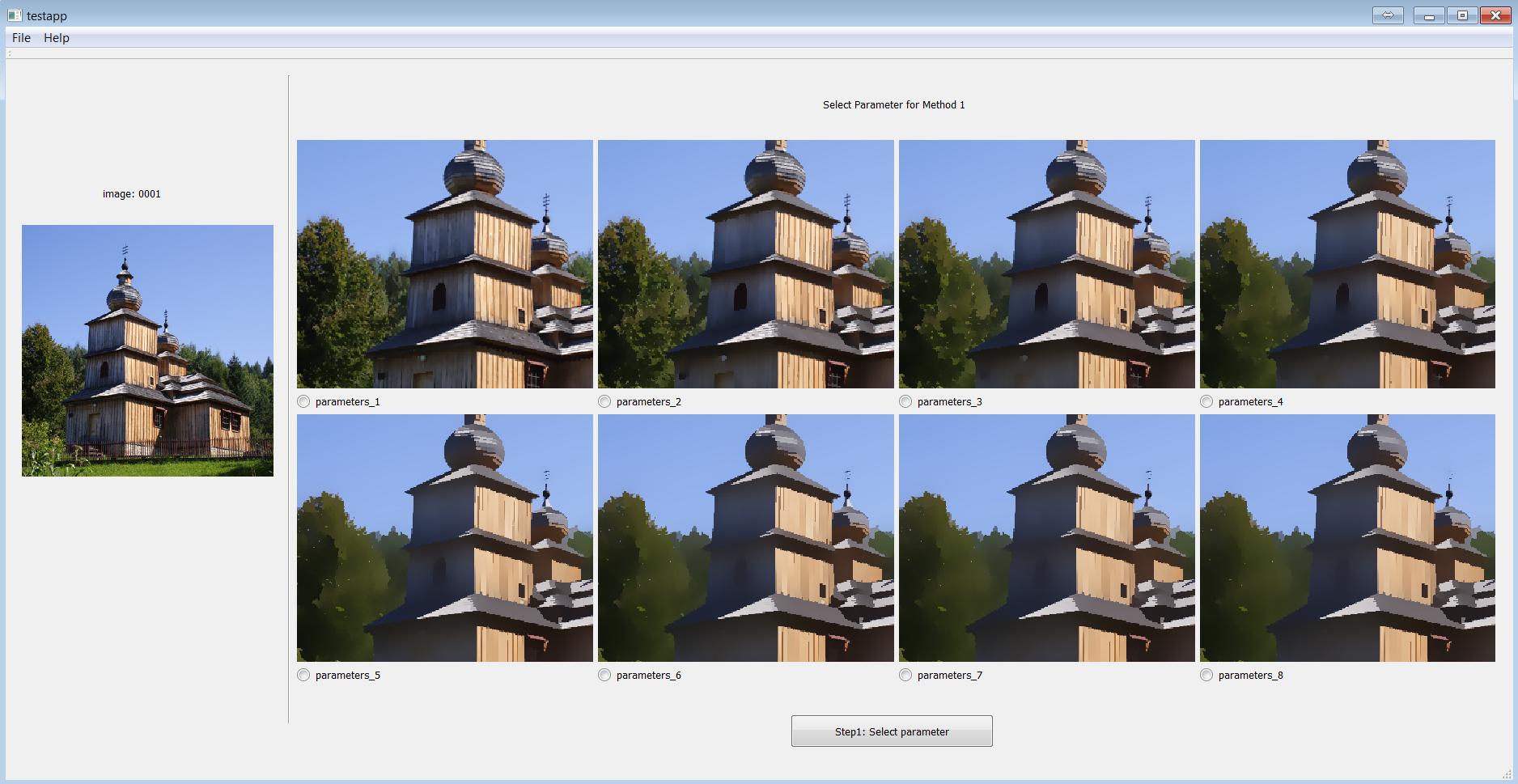}}\\
\subfloat[Step 2: Choose the best result from 7 algorithms]{\includegraphics[width=0.48\textwidth]{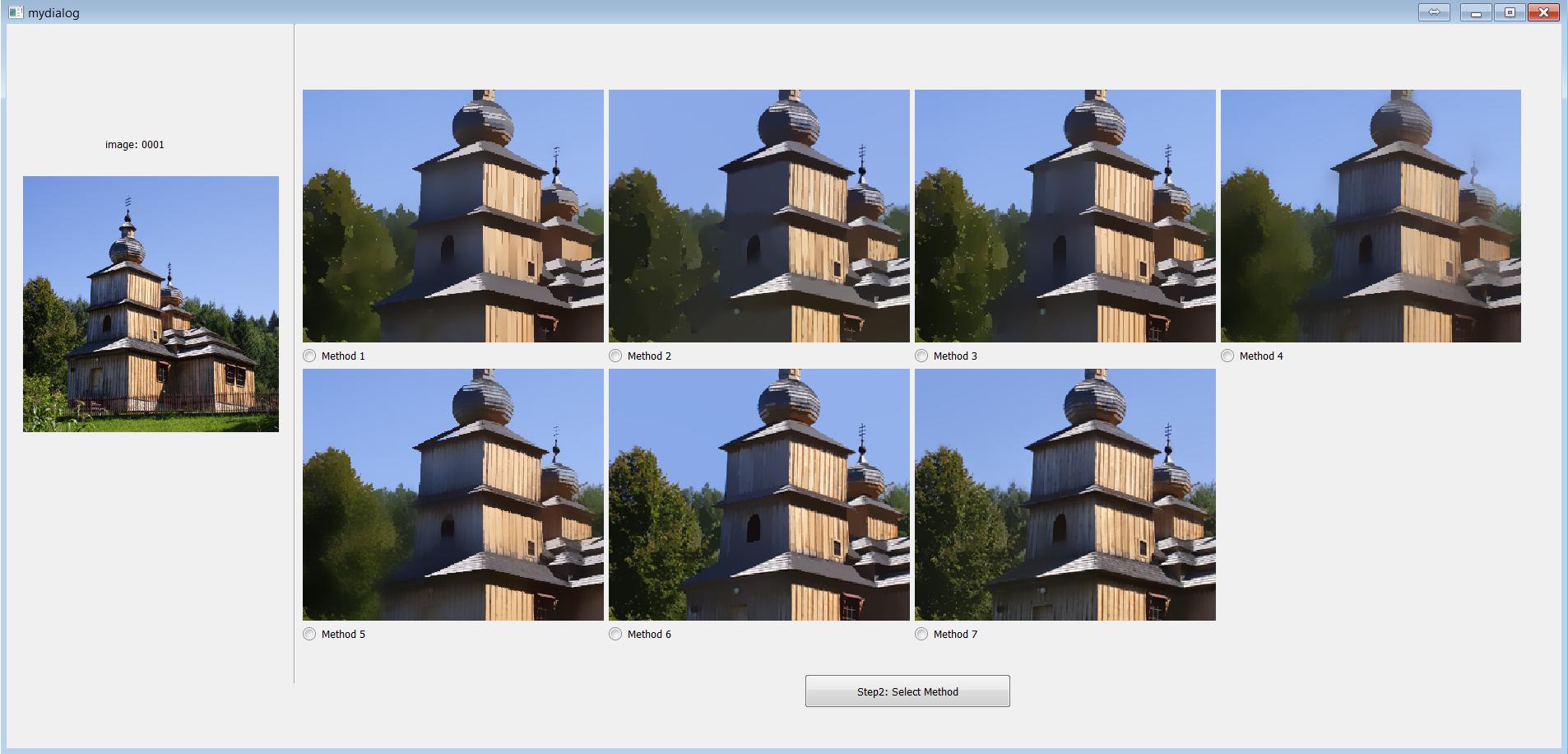}}
\caption{Snapshots of our two-step selection interface.}
\label{Fig:tool snapshot}
\end{figure}

\begin{table}[!t]
\centering
\caption{We predefined 8 sets of parameters for each smoothing algorithm. Additional parameters are set to default values suggested by the original authors. }
\label{table: parameter}
\resizebox{0.49\textwidth}{!}
{
\renewcommand{\arraystretch}{1.5}
\begin{tabular}{|l|l|l|l|l|l|l|l|l|l|}
\hline
\multicolumn{2}{|l|}{Parameters}                   & 1     & 2    & 3     & 4     & 5    & 6    & 7    & 8    \\ \hline
SD filter                             & $\lambda$  & 1     & 5    & 15    & 30    & 50   & 70   & 90   & 110  \\ \hline
\multicolumn{1}{|c|}{$L_0$ smoothing} & $\lambda$  & 0.005 & 0.01 & 0.02  & 0.03  & 0.04 & 0.05 & 0.06 & 0.08 \\ \hline
\multirow{2}{*}{FGS}                  & $\sigma_c$ & 0.02  & 0.02 & 0.025 & 0.025 & 0.03 & 0.03 & 0.04 & 0.04 \\ \cline{2-10}
                                      & $\lambda$  & 400   & 900  & 600   & 900   & 500  & 900  & 400  & 1200 \\ \hline
\multirow{2}{*}{Tree Filtering}       & $\sigma$   & 0.05  & 0.05 & 0.1   & 0.1   & 0.2  & 0.2  & 0.4  & 0.4  \\ \cline{2-10}
                                      & $\sigma_s$ & 8     & 4    & 8     & 4     & 8    & 4    & 8    & 4    \\ \hline
WMF                                   & $\sigma$   & 10    & 30   & 50    & 70    & 90   & 110  & 130  & 150  \\ \hline
\multirow{2}{*}{$L_1$ smoothing}      & $\alpha$   & 10    & 10   & 20    & 20    & 100  & 100  & 200  & 200  \\ \cline{2-10}
                                      & $\theta$   & 200   & 50   & 200   & 50    & 200  & 50   & 200  & 50   \\ \hline
\multirow{2}{*}{LLF}                  & $\sigma_r$ & 0.1   & 0.1  & 0.2   & 0.2   & 0.4  & 0.4  & 0.6  & 0.6  \\ \cline{2-10}
                                      & $\alpha$   & 2     & 4    & 2     & 4     & 2    & 4    & 2    & 4    \\ \hline
\end{tabular}
}
\end{table}

According to the fact that the best smoothing results of different images may come from different algorithms with different parameter settings, we have developed an interactive interface where one can choose the proper edge-preserving smoothing result in two steps:

$\bullet$ Step 1: Given a source image, choose a parameter setting for each algorithm which generates the best smoothing result for that algorithm .

$\bullet$ Step 2: Choose the best one from the best results of the seven algorithms.

Figure \ref{Fig:tool snapshot} shows snapshots of the selection tool. The source image is shown in the top left corner. We pre-defined eight parameter settings for each algorithm, as shown in Table \ref{table: parameter}. These eight settings are selected to cover a wide range of coarse-to-fine smoothing levels. For each algorithm, the smoothing results corresponding to these eight parameter settings are presented on the same screen (Figure \ref{Fig:tool snapshot}(a)). A user first chooses the best result produced by different parameters for each method. Afterwards, the seven smoothing results chosen from the previous step, one for each of the seven algorithms, are shown side by side on a pop-out window (Figure \ref{Fig:tool snapshot}(b)). The user then chooses the best result from the seven as the final selection for the current source image.

Since the above two steps heavily rely on visual comparisons and multiple images need to be shown on the same screen, we use a 32-inch Truecolor IPS monitor with a high resolution ($3840 \times 2160$) for dataset construction.

\subsection{Selection Protocol}
%\old{Before going any further, let's discuss the criteria of edge-preserving smoothing. As is widely acknowledged, the general criterion for basic edge-preserving smoothing is that the visually salient edges should be preserved and other textural regions should be flattened. We aim to explicitly capture this general criterion by asking several subjects to conduct visual inspection and selection, as will be discussed in the following. Beside the general criterion, for each related application, such as tone mapping, there may be some task-specific requirements for the filter. We implicitly capture this criterion by creating our ground truth images from the results by the selected seven filters, which are originally designed for a variety of applications. In Section \ref{sec6}, we show that such criterion is sufficient to obtain consistently good results in related applications.}

As is widely acknowledged, the general criterion for basic edge-preserving smoothing is that the visually salient edges of major structures should be preserved while the trivial details should be removed. Subjects are instructed to select good edge-preserving smoothing results from different edge-preserving smoothing (EPS) filters with different parameter settings. 

Since humans could have different perceptual comprehension of trivial details and major structures, they might select different smoothed images as their preferred edge-preserving smoothing results. To model such perceptual differences among human users, we formed a subject group of 26 volunteers, all of whom are graduate students in The University of Hong Kong and The Hong Kong Polytechnic University. Each source image in our dataset and its associated smoothing results were randomly assigned to 14 volunteers. That is, on average one volunteer was asked to select proper smoothing images for 267 source images. Most of the volunteers do not have prior experiences on edge-preserving image smoothing. Simple but key instructions were given to the volunteers in order to familiarize them with the nature of edge-preserving smoothing.

$\bullet$ Instruction 1: Strong edges should be preserved and blurry effects at significant edges are extremely undesired.

$\bullet$ Instruction 2: The color of a smoothed image should be as close to the original image as possible.

$\bullet$ Instruction 3: Under instructions 1 and 2, the smoother, the better.

The volunteers were further presented with a few unambiguous examples of edge-preserving smoothing for training. It typically takes a volunteer around 2 minutes to select one final result. In order to minimize the negative effects of visual fatigue, only a single work session up to 60 minutes is allowed in any single day.

In addition, many types of EPS algorithms have been proposed or tailored to meet specific criterion of their targeted applications (e.g., tone mapping). In our proposed benchmark construction, we carefully selected 7 representative EPS algorithms, which are designed for various applications, to create the ground truths. The constructed dataset implicitly captures various good edge-preserving smoothing properties. As a consequence, the benchmark is applicable to a wide range of applications. We will demonstrate the effectiveness of our benchmark in applications of tone mapping and contrast enhancement in Section \ref{sec6}.

\subsection{Dataset Statistics}
\label{Dataset Statistics}

\begin{figure}[!t]
%\captionsetup[subfigure]{subrefformat=simple, listofformat=subsimple, labelformat=empty}
\centering
\subfloat[Vote distribution among different algorithms]{\includegraphics[width=0.49\textwidth]{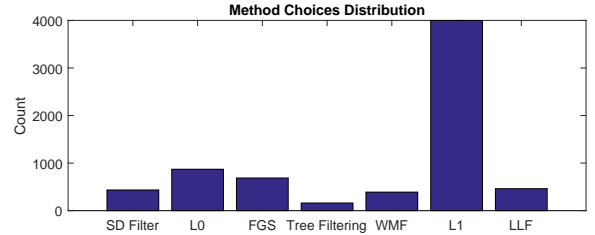}}\\
\subfloat[Vote distribution among different parameter settings of $L_1$ smoothing ]{\includegraphics[width=0.49\textwidth]{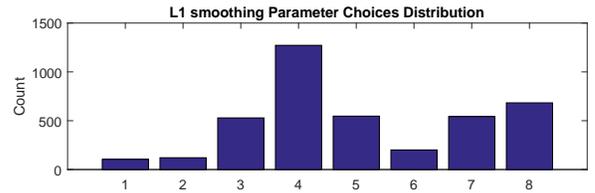}}
\caption{Distributions of human selection results. Two distributions of users' votes over different smoothing algorithms and over different parameter settings of the algorithms of $L_1$ smoothing.}
\label{Fig:distribution}
\end{figure}

The constructed dataset for edge-preserving smoothing contains 500 natural images (400 for training and 100 for testing). As mentioned above, to reduce the bias of subject preference during manual selection, we collected 14 human-selected smoothing results for each image. The entire process lasted for one and half months.

Since each image is finally associated with 14 human-selected smoothing results, there are 7000 choices in total. As shown in Figure \ref{Fig:distribution}(a), a large proportion of the choices (3999 choices) are generated by the $L_1$ smoothing algorithm. Nevertheless, there are still a considerable number of choices (3001) distributed among the other 6 smoothing algorithms.  Figure~\ref{Fig:distribution}(b) shows the distribution of the parameter choices for the $L_1$ smoothing method. This distribution confirms the observation that the proper smoothing result of different images may come from different algorithms with different parameter settings.

\begin{figure}[!t]
\centering
\includegraphics[width=0.49\textwidth]{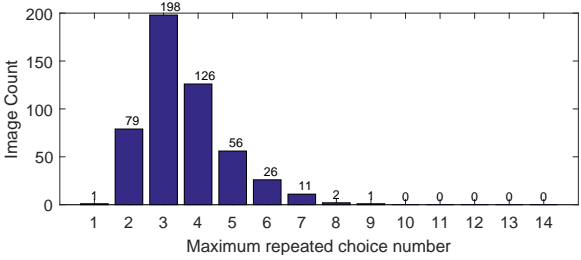}
\caption{Distribution of the maximum number of repeated choices.
}
\label{Fig:repeat_number}
\end{figure}

To show the consistency of choices of different volunteers on the same source image, we compute the maximum number of repeated choices for each image. Let ${count_t(m,p)}$ denote the number of volunteers who chose method $m$ with parameter setting $p$ to compute the proper smoothing result of the $t$-{th} source image. Note that $\sum_{m=1}^{7} \sum_{p=1}^{8} count_t(m,p)=14$. The maximum number of repeated choices for image $t$ is defined as $\max_{m} \max_{p} count_t(m,p)$. The distribution of the maximum number of repeated choices across all images is shown in Figure~\ref{Fig:repeat_number}. We can see that there are 420 (out of 500) images whose maximum number of repeated choices is greater than or equal to 3. In other words, for 84\% of the source images, at least 3 volunteers chose the result from the same algorithm with the same parameter.

\section{Quantitative Measures}
\label{Evaluation Measure}

Denote by $x^{t}_{i,j}$ the pixel value at position $(i,j)$ of the $t$-th source image in our dataset and denote by $y^{t,k}_{i,j}$ the pixel value of the corresponding ``groundtruth" smoothed image selected by the $k$-th volunteer ($k \in [1,2,...,14]$) .
%Given an edge-preserving filter $F$, we let $F(x^t)$ denote the smoothed $t$-th image by this filter.
We measure the quality of an edge-preserving filter $F$ in terms of the Root Mean Squared Error (RMSE) and Mean Absolute Error (MAE). In our problem, there are multiple ``groundtruth" smoothed images selected by different subjects, and we define the RMSE and MAE as follows:

\begin{equation}
\label{eq:rmse1}
    RMSE = \left(\frac{\sum_t \sum_{i,j} \sum_{k=1}^{14} \frac{1}{14} ||F(x^t)_{i,j}-y^{t,k}_{i,j}||^2}{\sum_t \sum_{i,j}}\right)^{\frac{1}{2}}
\end{equation}
\begin{equation}
\label{eq:mae1}
    MAE = \frac{\sum_t \sum_{i,j} \sum_{k=1}^{14} \frac{1}{14} ||F(x^t)_{i,j}-y^{t,k}_{i,j}||_1 }{\sum_t \sum_{i,j}}
\end{equation}
where $F(x^t)_{i,j}$ is the pixel value in the smoothed image produced by the edge-preserving filter $F$. The denominator $\sum_t \sum_{i,j}$ denotes the total number of pixels in all images.

Due to the subjective nature of image quality assessment, inevitably there exist noises and outliers in the ``groundtruth" smoothed images selected by different subjects. To reduce the effect of noises and outliers on performance evaluation, we take a voting strategy to focus on those smoothed images chosen by more subjects.

\begin{table}[!t]
\centering
\caption{The minimum WRMSE and WMAE of existing state-of-the-art edge-preserving smoothing methods and deep models. The optimal parameter setting of each algorithm is used across the entire dataset. {\color{red}Red}, {\color{green}Green} and {\color{blue}Blue} color indicates the best, second best and third best results, respectively. }
\label{table:method evaluation}
\resizebox{0.49\textwidth}{!}
{
\renewcommand{\arraystretch}{2}
\begin{tabular}{|l|l|l|l|l|l|l|l|l|l|}
\hline
Error     & SD filter & $L_0$ smooth & FGS   & TreeFilter & WMF   & $L_1$ smooth & LLF   & VDCNN & ResNet \\ \hline
$\mbox{WRMSE}^*$ & 11.57     & 10.64        & 10.67 & 14.31      & 11.83 & \color{blue}9.89        & 11.06 & \color{green}9.78  & \color{red}9.03   \\ \hline
$\mbox{WMAE}^*$ & 7.65      & 6.93         & 6.82  & 9.24       & 7.96  & \color{green}5.76         & 7.29  & \color{blue}6.15  & \color{red}5.55   \\ \hline
\end{tabular}
}
\end{table}

\begin{figure*}[!t]
\captionsetup[subfigure]
  {subrefformat=simple, listofformat=subsimple,farskip=1pt}
\centering
\subfloat[VDCNN]{\includegraphics[width=0.95\textwidth]{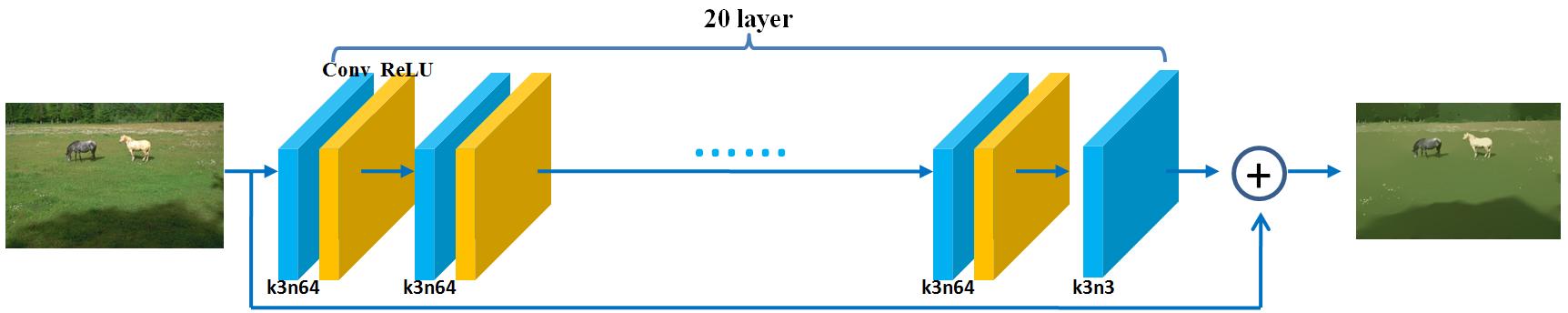}}\
\subfloat[ResNet]{\includegraphics[width=0.95\textwidth]{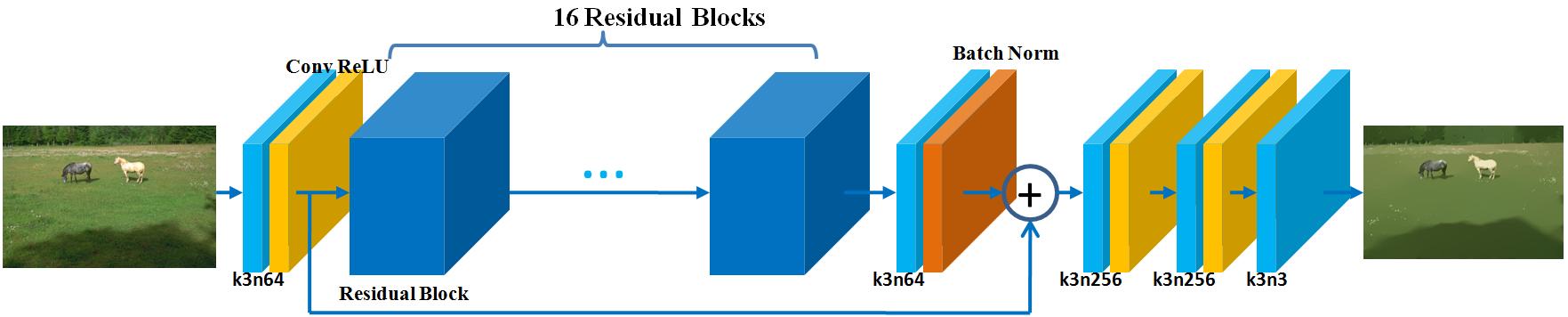}}
\caption{Network architecture of (a) VDCNN and (b) ResNet. Each convolutional layer is denoted with kernel size (k) and number of feature maps (n). The stride is 1 for all convolutional layers. Residual block is illustrated in Figure \ref{Fig:ResidualBlock}.}
\label{fig:architecture}
\end{figure*}

\begin{figure}[!t]
\centering
\includegraphics[width=0.4\textwidth]{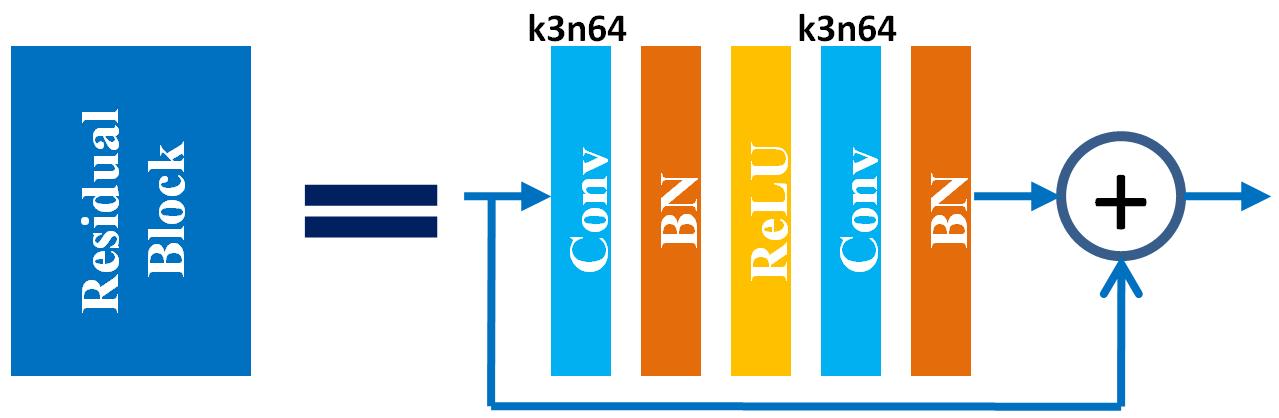}
\caption{Internal structure of a residual block used in ResNet.
}
\label{Fig:ResidualBlock}
\end{figure}

As mentioned in Section \ref{Dataset Statistics}, $count_t(m,p)$ denotes the number of subjects who chose method $m$ with parameter setting $p$ as the best smoothing filter of the $t$-th source image.  Each image is associated with 14 human-selected smoothing results, and there are 7000 choices in total. The total number of times that method $m$ with parameter setting $p$ was chosen by a subject is denoted by COUNT$(m,p)$.

The voting strategy is that for each source image, we sort its choice numbers ($count_t(m,p)$) in a descending order. If there is a tie between different combinations of methods and parameter settings, we sort them according to COUNT$(m,p)$. That is because the total number of times a method with one of its parameter settings was chosen is an indicator of its overall performance. We tend to choose the smoothed images produced by more reliable methods when user preferences are the same. We only keep the first five results ``groundtruth" smoothed images. For example, if method $m$ with parameter $p$ was selected by most subjects for the $t$-th source image, we let $Y^{t,1}$ denote the smoothed image produced by method $m$ and parameter $p$, and set $count(Y^{t,1})=count_t(m,p)$. We denote by $Y^{t,2}$ the second most frequently chosen smoothed image, and so on. The quantitative measures defined in Equations \ref{eq:rmse1}-\ref{eq:mae1} can be extended to the weighted RMSE (WRMSE) and weighted MAE (WMAE) as follows:
\begin{equation}
\label{eq:rmse2}
    WRMSE = \left(\frac{\sum_t \sum_{i,j} \sum_{k=1}^{5} w_{t,k} ||F(x^t)_{i,j}-Y^{t,k}_{i,j}||^2 }{\sum_t \sum_{i,j}}\right)^{\frac{1}{2}}
\end{equation}
\begin{equation}
\label{eq:mae2}
    WMAE = \frac{\sum_t \sum_{i,j} \sum_{k=1}^{5} w_{t,k} ||F(x^t)_{i,j}-Y^{t,k}_{i,j}||_1}{\sum_t \sum_{i,j}}
\end{equation}
where $w_{t,k}$ is defined as:

\begin{equation}
\label{eq:w_tk}
    w_{t,k}= \frac{count(Y^{t,k})}{\sum_{k=1}^{5} count(Y^{t,k})}
\end{equation}

In our quantitative measures, the weight of a combination of a method and a parameter setting varies across different source images because none of the existing smoothing algorithms performs consistently well over a wide range of image contents. The proposed quantitative measures assign higher weights to ``groundtruth" smoothed images chosen by more subjects while excluding noises and outliers at the same time. We use Equations \ref{eq:rmse2} and \ref{eq:mae2} to quantitatively measure the performance of various edge-preserving smoothing algorithms as well as our trained models in the rest of this paper.

\subsection{Evaluation of Existing Algorithms}
\label{Evaluation of Previous Method}

The parameter setting of a smoothing algorithm affects its performance. We measure WRMSE and WMAE across the entire testing set of our dataset with different parameter settings for each of the seven chosen algorithms. The minimum WRMSE and WMAE are denoted by $\mbox{WRMSE}^*$ and $\mbox{WMAE}^*$, respectively, and the optimal parameter setting of each method was determined by greedy search. The process is as follows: set a group of parameter settings for each algorithm; apply them to all testing images; record WRMSE and WMAE. The parameter settings that make the seven chosen methods achieve their $\mbox{WRMSE}^*$ are given as follows: SD filter ($\lambda=5$), $L_0$ smoothing ($\lambda$=0.02), FGS ($\sigma_c=0.025,\lambda=600$), Tree Filtering ($\sigma=0.05,\sigma_s=8$),  WMF ($\sigma$=30),  $L_1$ smoothing ($\alpha=20,\theta=50$), LLF ($\sigma_r=0.4,\alpha=2$). The parameter settings that make the seven methods achieve their $\mbox{WMAE}^*$ are given as follows: SD filter ($\lambda=5$), $L_0$ smoothing ($\lambda$=0.01), FGS ($\sigma_c=0.025,\lambda=600$), Tree Filtering ($\sigma=0.05,\sigma_s=8$),  WMF ($\sigma$=50),  $L_1$ smoothing ($\alpha=20,\theta=50$), LLF ($\sigma_r=0.2,\alpha=4$). Additional parameters are set to default values suggested by original authors.

From Table~\ref{table:method evaluation}, we can see that the $L_1$ smoothing algorithm has lower $\mbox{WRMSE}^*$ and $\mbox{WMAE}^*$ than other smoothing algorithms because the results generated by the $L_1$ smoothing algorithm were most frequently chosen by the volunteers as their preferred results when our dataset was constructed, as shown in Figure~\ref{Fig:distribution}.

\section{Deep Learning Models}

Deep neural networks have achieved great successes in low-level computer vision problems, including reproducing edge-preserving filters \cite{xu2015deep,liu2016learning,li2016deep}, image denoising \cite{mao2016imagedenoise,zhang2017beyond}, image super-resolution \cite{kim2016accurate,ledig2016photo-realistic,kim2016deeply,Tai-DRRN-2017,Tai-MemNet-2017}, and JPEG deblocking \cite{dong2015compression,zhang2017beyond}.
To build the baseline models in our benchmark, we resort to the latest deep neural networks as learning-based baseline algorithms. Deep neural networks have a large number of parameters (weights), which can be optimized to address a specific task. A well-trained deep neural network on our dataset is expected to be able to produce high-quality results for a wide range of inputs. Thus, it is not necessary to tune parameters of the trained model for a new image, which is a desirable property for edge-preserving image smoothing.

%because recently deep fully convolutional networks with high-resolution output have had many successes in low-level computer vision~\cite{mao2016imagedenoise,kim2016accurate,kim2016deeply,Tai-DRRN-2017,Tai-MemNet-2017,zhang2017beyond} and such network architectures can typically be adapted for different low-level vision tasks.

%There have been many attempts to handle low-level image processing problems by deep neural networks. One representative network architecture is the very deep convolution network network (VDCNN), which can exploit high nonlinearities and model complex functions \cite{simonyan2014very} but also may meet vanishing/exploding gradients problem \cite{bengio1994learning}. Kim \etal \cite{kim2016accurate} used a 20-layer model to address the single image super-resolution (SISR) problem. Residual-learning, high learning rates and gradient clipping schemes are adopted to achieve fast convergence during training. Zhang \etal \cite{zhang2017beyond} incorporated Batch Normalization \cite{ioffe2015batch} and further extended it to Gaussian denoising, JPEG image deblocking tasks both achieving competitive performance compared to state-of-the-art methods.

%Another network architecture, deep residual network with skip-connections (ResNet) \cite{ledig2016photo-realistic,lim2017enhanced}, is mentioned here since it sets the new state-of-the-art for image super resolution.

In this section, we present two representative network architectures and report their performance as a baseline for our edge-preserving smoothing dataset. Specifically, we employ representative deep convolutional neural network (CNN) architectures for our problem. This is because recently, deep CNNs have been successfully used in many low-level vision tasks~\cite{mao2016imagedenoise,kim2016accurate,kim2016deeply,Tai-DRRN-2017,Tai-MemNet-2017,zhang2017beyond} and their network architectures can be employed for different tasks, including edge-preserving smoothing.

\subsection{Network Architecture}
\noindent \textbf{VDCNN:} 20 layers are stacked to form a very deep convolutional neural network, as shown in Figure \ref{fig:architecture}(a). The layers maintain the same spatial resolution, the same kernel size ($3 \times 3$) and the same number of feature maps (64) except the last output layer which has 3 channels only. The original model proposed by Kim \textit{et al}.~\cite{kim2016accurate}  works on the luminance channel of an image, which is the common practice in the literature of single image super-resolution. In contrast, we perform the training on three RGB channels since all color channels change during edge-preserving smoothing.

\noindent \textbf{ResNet:} Residual networks~\cite{ledig2016photo-realistic,lim2017enhanced,he2016deep} have exhibited outstanding performance in both low-level and high-level computer vision problems. As shown in Figure \ref{Fig:ResidualBlock}, a basic residual block includes convolutional layers (Conv), batch normalization (BN), rectified linear units (ReLU) and a skip connection. A complete ResNet architecture is shown in Figure \ref{fig:architecture}(b). It was originally proposed in \cite{ledig2016photo-realistic}, where it is used for inferring photo-realistic high-resolution images from low-resolution ones. We replace ParametricReLU \cite{he2015delving} by ReLU and remove the up-sampling layer here.

\begin{figure}[!t]
    \captionsetup[subfigure]
    {subrefformat=simple, listofformat=subsimple,farskip=1pt,labelformat=empty}
    \centering
    \subfloat[Original images]{\includegraphics[width=0.22\textwidth]{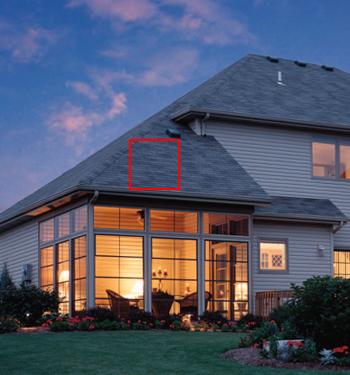}}
    \hspace{-0.11\textwidth}\subfloat{\includegraphics[width=0.11\textwidth]{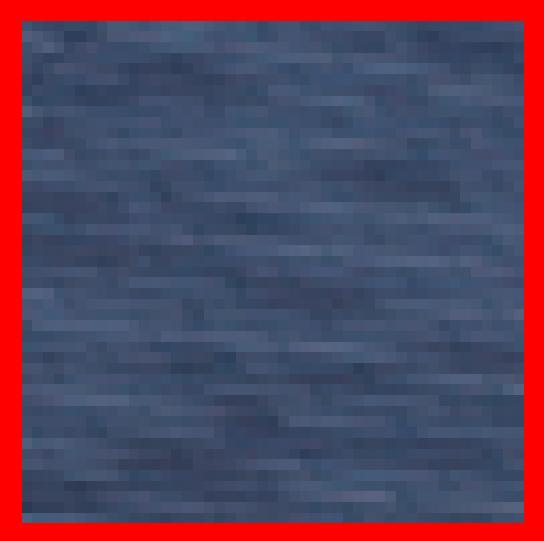}}~
    \subfloat[$loss_{l_2}$]{\includegraphics[width=0.22\textwidth]{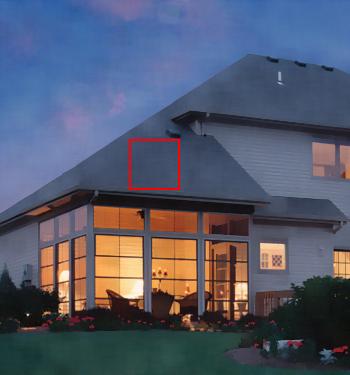}}
    \hspace{-0.11\textwidth}\subfloat{\includegraphics[width=0.11\textwidth]{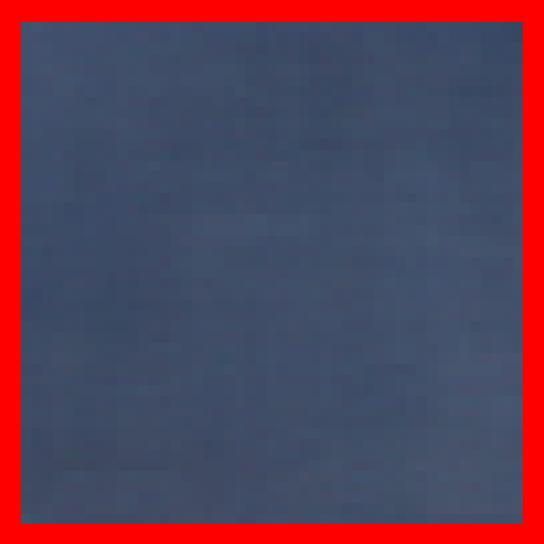}}\
    \subfloat[$loss_{l_1}$]{\includegraphics[width=0.22\textwidth]{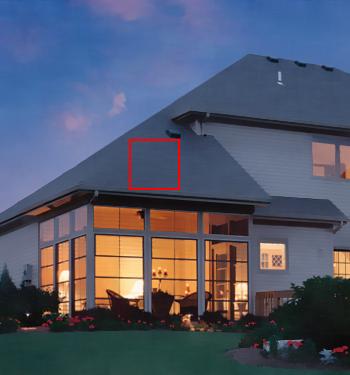}}
    \hspace{-0.11\textwidth}\subfloat{\includegraphics[width=0.11\textwidth]{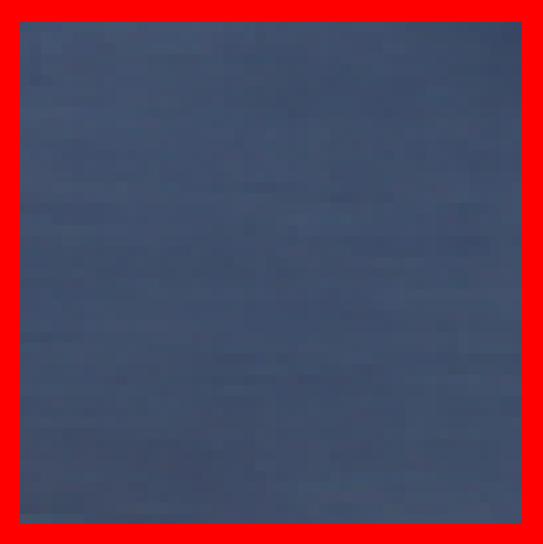}}~
    \subfloat[$loss_{l_1}+loss_{nb}$]{\includegraphics[width=0.22\textwidth]{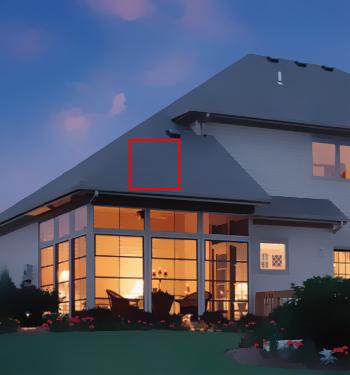}
        \hspace{-0.115\textwidth}\subfloat{\includegraphics[width=0.11\textwidth]{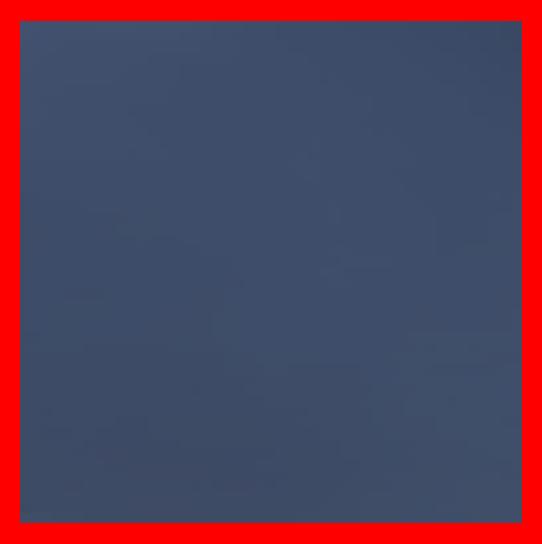}}}
    \caption{Example of smoothing outputs using different losses. We can see that the deep model trained with $loss_{l_1}+loss_{nb}$ produces smoother result at sky, roof and grass regions than $loss_{l_2}$ or $loss_{l_1}$ alone.}
    \label{fig:l1 l2 loss}
\end{figure}

\subsection{Loss Functions}

Since there exist multiple ``groundtruth" smoothed images for each source image in our dataset, we define a weighted $L_2$ loss (Equation \ref{eq:l2 loss}) and a weighted $L_1$ loss (Equation \ref{eq:l1 loss}) in a manner similar to the weighted RMSE in Equation \ref{eq:rmse2} and the weighted MAE in Equation \ref{eq:mae2}:

\begin{equation}
\label{eq:l2 loss}
    loss_{l_2} = \sum_t \sum_{i,j} \sum_{k=1}^{5} w_{t,k} \left\|M_{\theta}(x^t)_{i,j}-Y^{t,k}_{i,j}\right\|^2,
\end{equation}
\begin{equation}
\label{eq:l1 loss}
    loss_{l_1}  = \sum_t \sum_{i,j} \sum_{k=1}^{5} w_{t,k} \left\|M_{\theta}(x^t)_{i,j}-Y^{t,k}_{i,j}\right\|_1,
\end{equation}
where $M_{\theta}(x^t)$ represents the output from a deep network and $x^t$ is the input image.

%\begin{figure*}[!t]
%	\captionsetup[subfigure]
%	{subrefformat=simple, listofformat=subsimple,farskip=1pt,labelformat=empty,justification=centering}
%	\centering
%	\subfloat[original images]{\includegraphics[width=0.4\textwidth]{figure/last_figure/origin_470_matlab.jpg}}
%	\hspace{-0.15\textwidth}\subfloat{\includegraphics[width=0.15\textwidth]{figure/last_figure/origin_470_local.jpg}}
%	\hspace{0.1in}	
%	\subfloat[$L_1$ smoothing]{\includegraphics[width=0.4\textwidth]{figure/last_figure/l1_470_matlab.jpg}}
%	\hspace{-0.15\textwidth}\subfloat{\includegraphics[width=0.15\textwidth]{figure/last_figure/l1_470_local.jpg}}\\
%	\subfloat[VDCNN]{\includegraphics[width=0.4\textwidth]{figure/last_figure/vdcnn_470_matlab.jpg}}
%	\hspace{-0.15\textwidth}\subfloat{\includegraphics[width=0.15\textwidth]{figure/last_figure/vdcnn_470_local.jpg}}\hspace{0.1in}
%	\subfloat[ResNet]{\includegraphics[width=0.4\textwidth]{figure/last_figure/ResNet_470_matlab.jpg}}
%	\hspace{-0.15\textwidth}\subfloat{\includegraphics[width=0.15\textwidth]{figure/last_figure/ResNet_470_local.jpg}}
%	\caption{Comparison between $L_1$ smoothing algorithm and deep models.}
%	\label{fig:compare l1 ResNet}
%\end{figure*}

\begin{table}[!t]
\centering
\caption{Performance comparison between the two baseline network architectures under different loss functions.}
\label{table:loss}
\resizebox{0.48\textwidth}{!}
{
\renewcommand{\arraystretch}{1.5}
\begin{tabular}{|l|l|l|c|l|l|c|}
\hline
\multirow{2}{*}{Error} & \multicolumn{3}{c|}{VDCNN}                                         & \multicolumn{3}{c|}{ResNet}                                        \\ \cline{2-7}
                       & $loss_{l_2}$ & $loss_{l_1}$ & \multicolumn{1}{l|}{$loss_{l_1}+loss_{nb}$} & $loss_{l_2}$ & $loss_{l_1}$ & \multicolumn{1}{l|}{$loss_{l_1}+loss_{nb}$} \\ \hline
WRMSE                   & 10.14   & 9.90   & 9.78                                           & 9.58    & 9.51    & 9.03                                           \\ \hline
WMAE                    & 6.92    & 6.20    & 6.15                                           & 6.50    & 6.12    & 5.55                                          \\ \hline
\end{tabular}
}
\end{table}

In addition to the weighted $L_2$ loss and weighted $L_1$ loss which enforce the consistency between the predicted smoothed images and their corresponding groundtruth smoothed images, we propose a neighborhood loss (Equation \ref{eq:Regu}) to explicitly encourage a network to learn the local variations of the groundtruth images:

\begin{equation}
\label{eq:Regu}
    \begin{split}
     &loss_{nb} = \sum_t \sum_{i,j} \sum_{k=1}^{5} \sum_{(p,q)\in N_{i,j}} \\
     &~~~~~~w_{t,k} \left\|(M_{\theta}(x^t)_{i,j}-M_{\theta}(x^t)_{p,q})-(Y^{t,k}_{i,j}-Y^{t,k}_{p,q})\right\|_1,
    \end{split}
\end{equation}
where $N_{i,j}$ denotes the $5\times5$ neightborhood centered at pixel $(i,j)$. The neighborhood term $||(M_{\theta}(x^t)_{i,j}-M_{\theta}(x^t)_{p,q})-(Y^{t,k}_{i,j}-Y^{t,k}_{p,q})||_1$ explicitly penalizes deviations in the gradient domain. We add this neighborhood term to the weighted $L_1$ loss since edge-preserving smoothing involves evident gradient changes.

Quantitative results are presented in Table~\ref{table:loss}, where we can see that ResNet achieves better performance than VDCNN when the same loss is used. $loss_{l_1}+loss_{nb}$ achieves better performance than $loss_{l_1}$ or $loss_{l_2}$ alone when the same network architecture is used, which validates the effectiveness of $loss_{nb}$. Figure~\ref{fig:l1 l2 loss} shows an example where we can see that the VDCNN trained with $loss_{l_1}+loss_{nb}$ produces smoother result at sky, roof and grass regions than $loss_{l_2}$ or $loss_{l_1}$ alone.

We thus take the VDCNN model and ResNet model trained with $loss_{l_1}+loss_{nb}$ as our baseline algorithms for our edge-preserving smoothing benchmark.

%Comparing (Resnet$+$weighted $L_1$ loss) and (ResNet$+$weighted $L_2$ loss), we find ResNet achieves lower WRMSE when trained with the weighted $L_2$ loss and lower WMAE when trained with the weighted $L_1$ loss. This is reasonable as the weighted $L_2$ loss and WRMSE are similar in definition while the weighted $L_1$ loss and WMAE are similar in definition.

%\old{Figure~\ref{fig:l1 l2 loss} shows an example where there are splotchy artifacts between the branches when the weighted $L_2$ loss is used. This is because the weighted $L_2$ loss penalizes large errors while being more tolerant to small errors. The weighted $L_1$ loss penalizes both large and small errors in a more uniform way, resulting in flatter regions.}

\begin{figure*}[!t]
\captionsetup[subfigure]
{subrefformat=simple, listofformat=subsimple,farskip=3pt}
\centering
\subfloat{\includegraphics[width=0.155\textwidth]{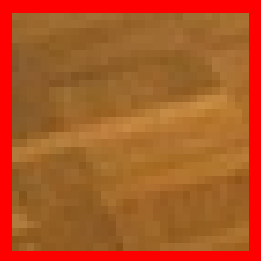}}\hspace{0.001in}
\subfloat{\includegraphics[width=0.156\textwidth]{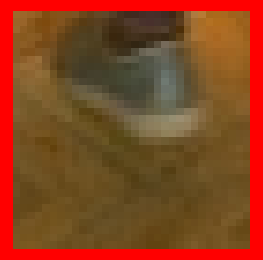}}~
\subfloat{\includegraphics[width=0.155\textwidth]{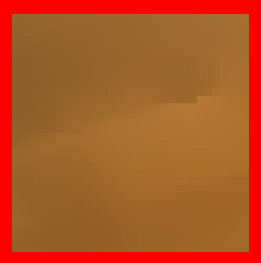}}\hspace{0.001in}
\subfloat{\includegraphics[width=0.156\textwidth]{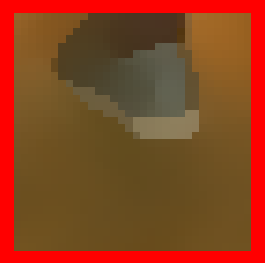}}~
\subfloat{\includegraphics[width=0.156\textwidth]{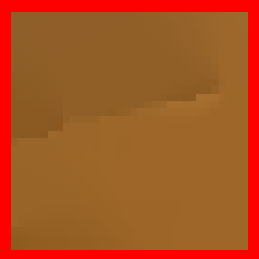}}\hspace{0.001in}
\subfloat{\includegraphics[width=0.155\textwidth]{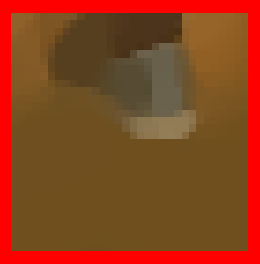}}\\
\addtocounter{subfigure}{-6}
\subfloat[\textbf{Source Image}]{\includegraphics[width=0.32\textwidth]{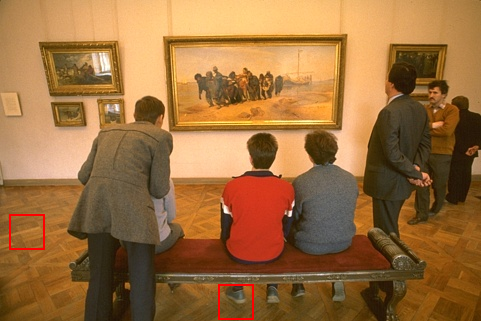}}~
\subfloat[\textbf{SD filter}]{\includegraphics[width=0.32\textwidth]{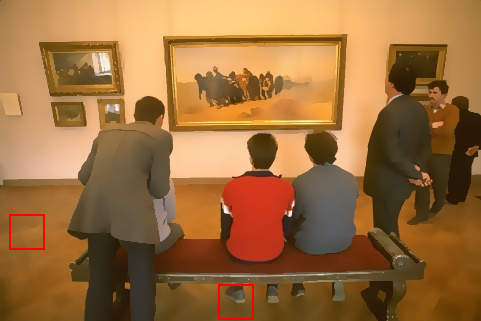}}~
\subfloat[\textbf{$L_0$ smoothing}]{\includegraphics[width=0.32\textwidth]{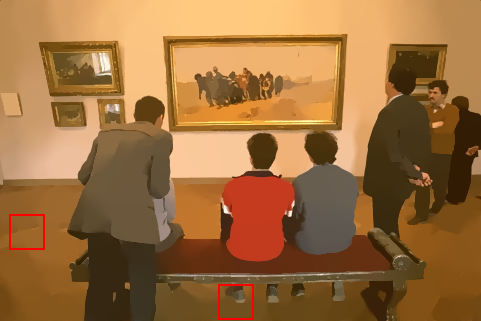}}\\
\subfloat{\includegraphics[width=0.155\textwidth]{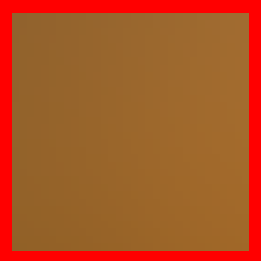}}\hspace{0.001in}
\subfloat{\includegraphics[width=0.155\textwidth]{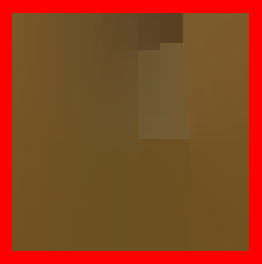}}~
\subfloat{\includegraphics[width=0.155\textwidth]{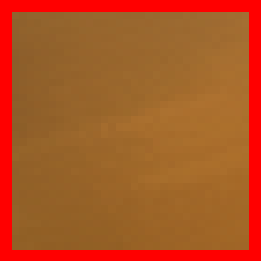}}\hspace{0.001in}
\subfloat{\includegraphics[width=0.155\textwidth]{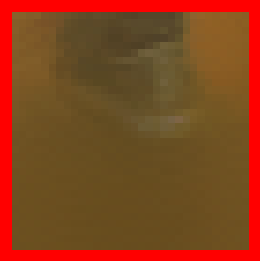}}~
\subfloat{\includegraphics[width=0.155\textwidth]{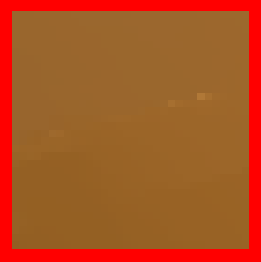}}\hspace{0.001in}
\subfloat{\includegraphics[width=0.155\textwidth]{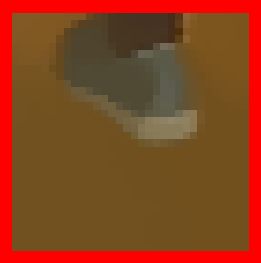}}\\
\addtocounter{subfigure}{-6}
\subfloat[\textbf{FGS}]{\includegraphics[width=0.32\textwidth]{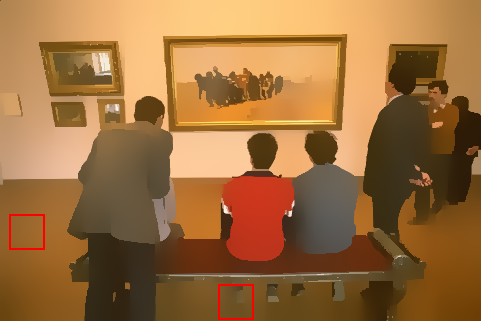}}~
\subfloat[\textbf{WMF}]{\includegraphics[width=0.32\textwidth]{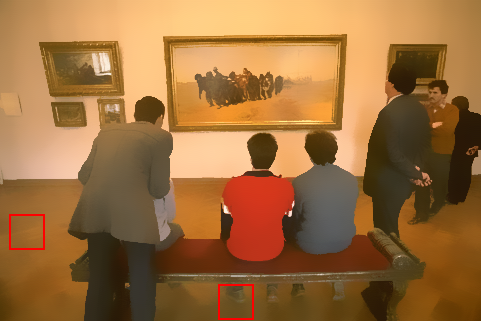}}~
\subfloat[\textbf{$L_1$ smoothing}]{\includegraphics[width=0.32\textwidth]{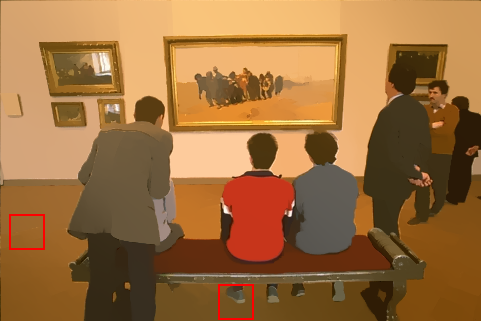}}\\
\subfloat{\includegraphics[width=0.1565\textwidth]{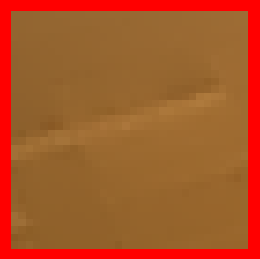}}\hspace{0.001in}
\subfloat{\includegraphics[width=0.155\textwidth]{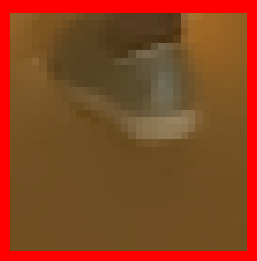}}~
\subfloat{\includegraphics[width=0.155\textwidth]{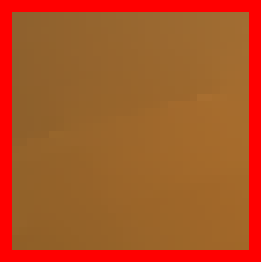}}\hspace{0.001in}
\subfloat{\includegraphics[width=0.155\textwidth]{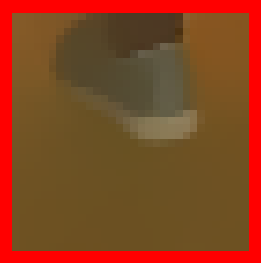}}~
\subfloat{\includegraphics[width=0.155\textwidth]{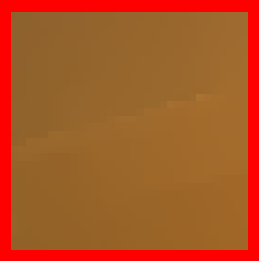}}\hspace{0.001in}
\subfloat{\includegraphics[width=0.155\textwidth]{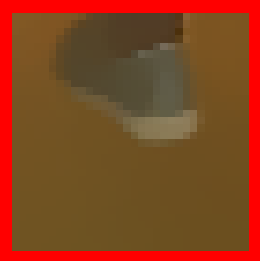}}\\
\addtocounter{subfigure}{-6}
\subfloat[\textbf{LLF}]{\includegraphics[width=0.32\textwidth]{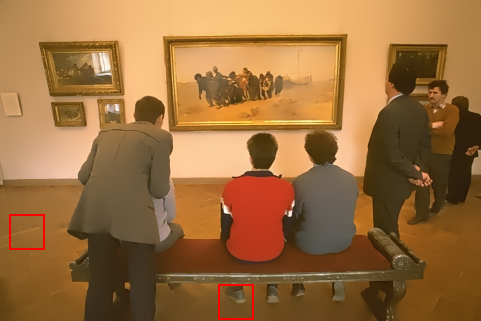}}~
\subfloat[\textbf{VDCNN}]{\includegraphics[width=0.32\textwidth]{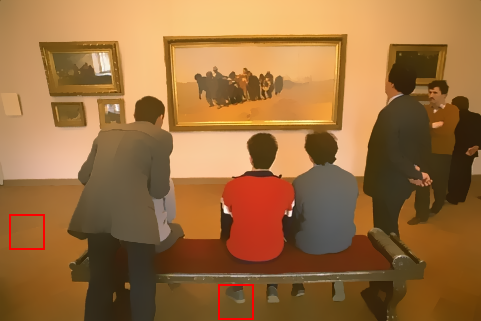}}~
\subfloat[\textbf{ResNet}]{\includegraphics[width=0.32\textwidth]{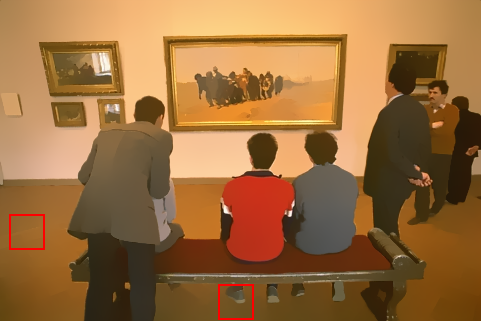}}
\caption{Comparison of edge-preserving smoothing results by existing state-of-the-art algorithms and deep models. (a) Source Image. (b-g) Results by SD filter, $L_0$ smoothing, FGS, WMF, $L_1$ smoothing and LLF, respectively. The parameters are set as the optimal parameters for $\mbox{WMAE}^*$ illustrated in Section \ref{Evaluation of Previous Method}. (h) VDCNN with $loss_{l_1}+loss_{nb}$. (i) ResNet with $loss_{l_1}+loss_{nb}$. }
    \label{fig:comparison model}
\end{figure*}

\begin{figure*}[!t]
\captionsetup[subfigure]
  {subrefformat=simple, listofformat=subsimple,farskip=1pt,labelformat=empty}
\centering
\subfloat{\includegraphics[width=0.23\textwidth]{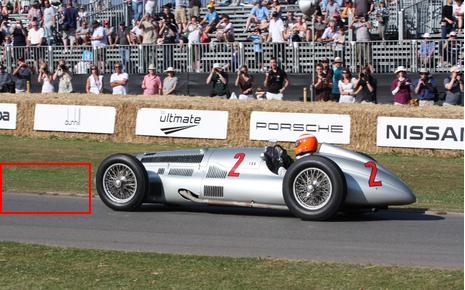}}~
\subfloat{\includegraphics[width=0.23\textwidth]{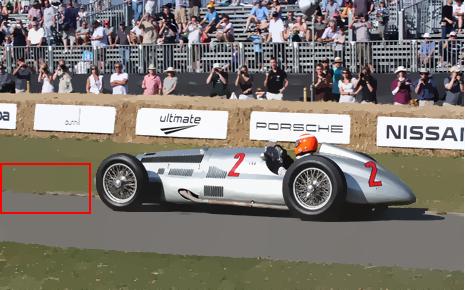}}~
\subfloat{\includegraphics[width=0.23\textwidth]{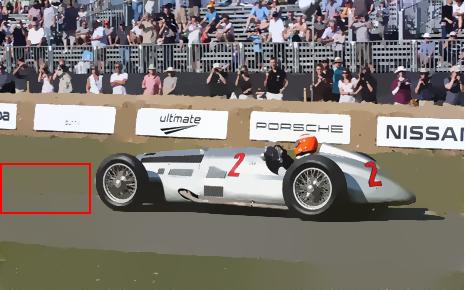}}~
\subfloat{\includegraphics[width=0.23\textwidth]{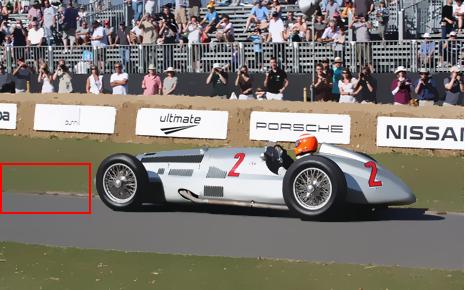}}\\
\vspace{-0.05in}
\subfloat{\includegraphics[width=0.23\textwidth]{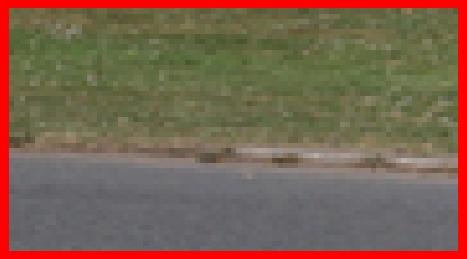}}~
\subfloat{\includegraphics[width=0.23\textwidth]{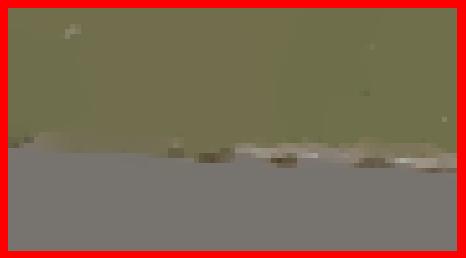}}~
\subfloat{\includegraphics[width=0.23\textwidth]{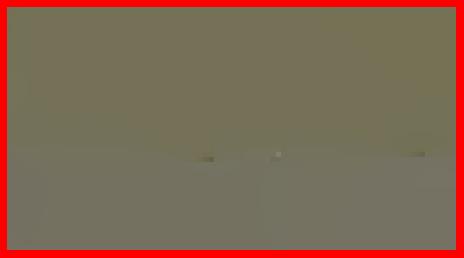}}~
\subfloat{\includegraphics[width=0.23\textwidth]{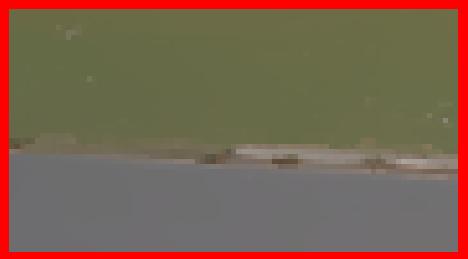}}\\
\vspace{0.05in}
%%%%%%%%%%%%%
\subfloat[Original images]{\includegraphics[width=0.23\textwidth]{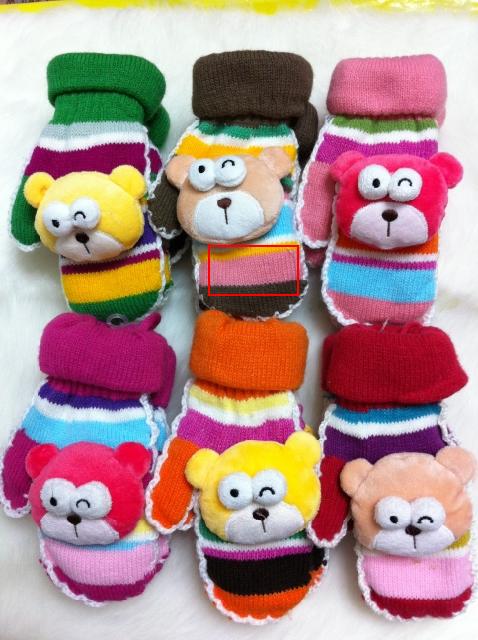}}
\hspace{-0.23\textwidth}\subfloat{\includegraphics[width=0.23\textwidth]{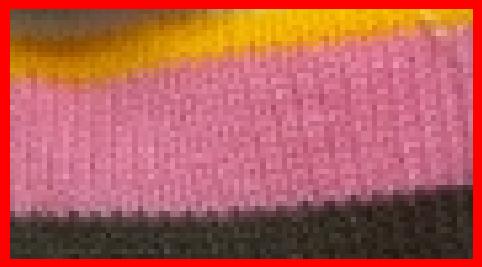}}~
\subfloat[$L_0$ smoothing($\lambda$=0.01)]{\includegraphics[width=0.23\textwidth]{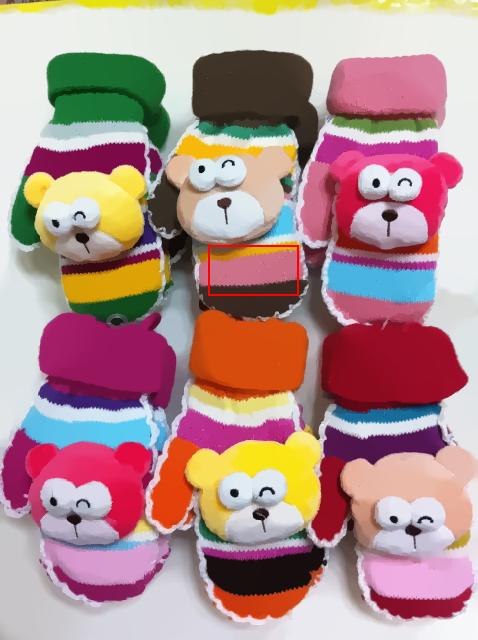}}
\hspace{-0.23\textwidth}\subfloat{\includegraphics[width=0.23\textwidth]{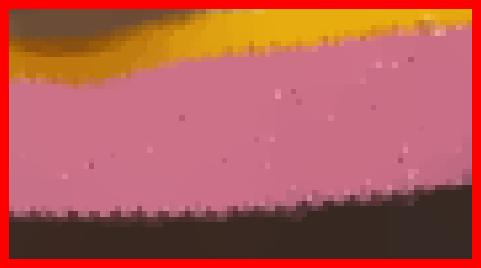}}~
\subfloat[$L_0$ smoothing($\lambda$=0.03)]{\includegraphics[width=0.23\textwidth]{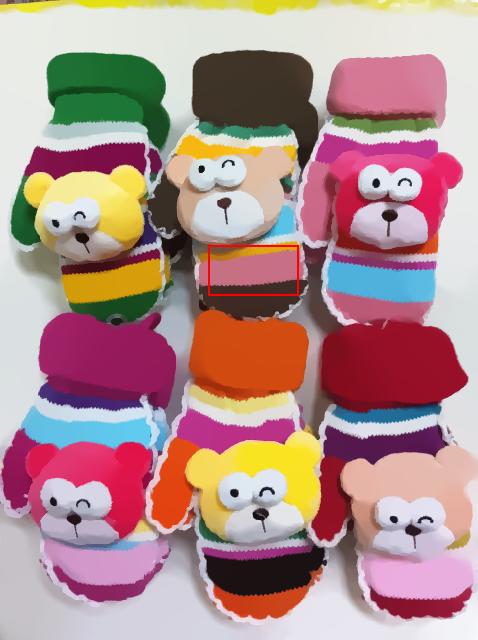}}
\hspace{-0.23\textwidth}\subfloat{\includegraphics[width=0.23\textwidth]{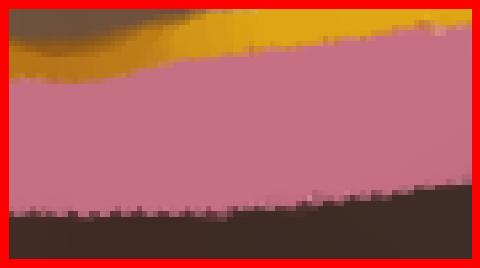}}~
\subfloat[ResNet]{\includegraphics[width=0.23\textwidth]{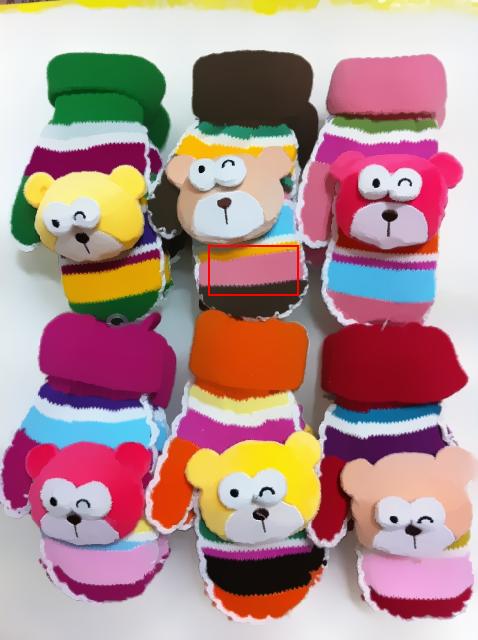}}
\hspace{-0.23\textwidth}\subfloat{\includegraphics[width=0.23\textwidth]{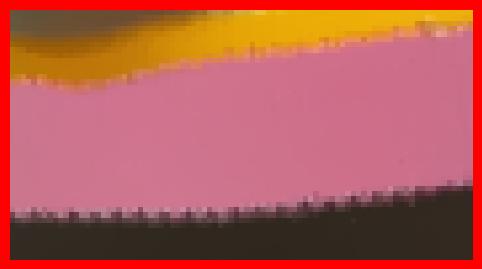}}\\
%%%%%%%%%%%
\caption{Comparison between $L_0$ smoothing algorithm and our ResNet model. $L_0$ smoothing algorithm needs different parameter settings for the 'Racing car' and the 'Gloves' images. If we set $\lambda$=0.03 for the 'Racing car' image, the edge between grass and road will blur. $\lambda$=0.01 is the proper setting. However, if we set $\lambda$=0.01 for the 'Gloves' image, there will be undesirable noises. In contrast, our ResNet model produces robust visual results on different images without tuning parameters.}
\label{fig:TuneParameter}
\end{figure*}

\begin{figure}[!t]
\captionsetup[subfigure]
  {subrefformat=simple, listofformat=subsimple,farskip=1pt,labelformat=empty}
\centering
\subfloat[origin images]{\includegraphics[width=0.24\textwidth]{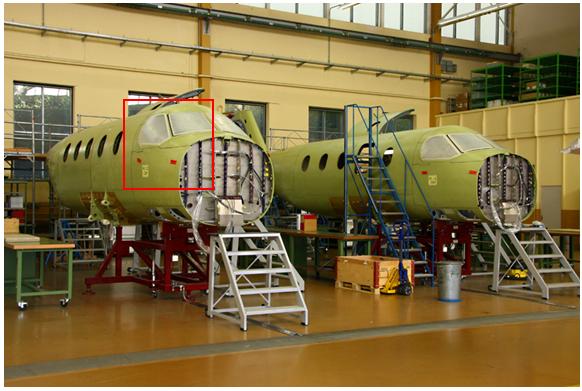}}
\hspace{-0.1\textwidth}\subfloat{\includegraphics[width=0.1\textwidth]{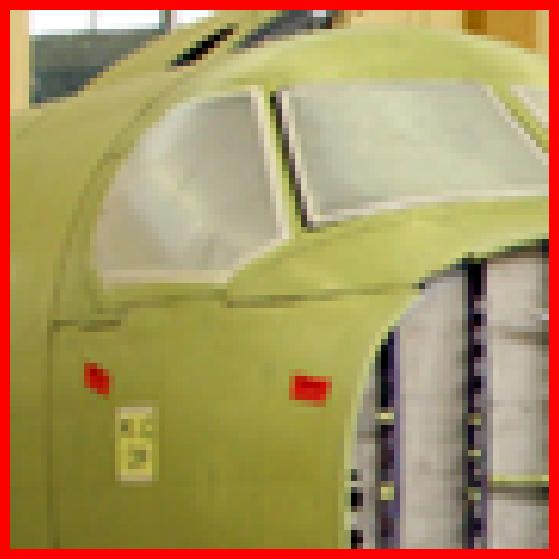}}
\hspace{0.001\textwidth}
\subfloat[$L_1$ smoothing]{\includegraphics[width=0.24\textwidth]{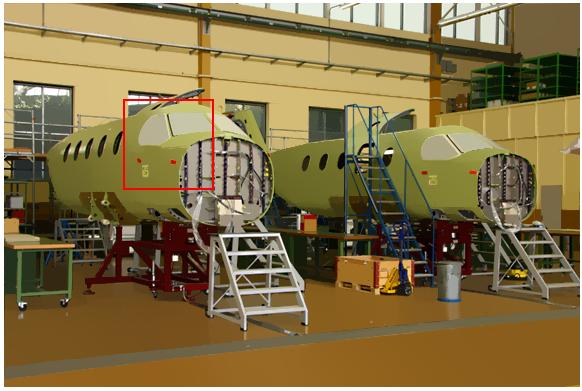}}
\hspace{-0.1\textwidth}\subfloat{\includegraphics[width=0.1\textwidth]{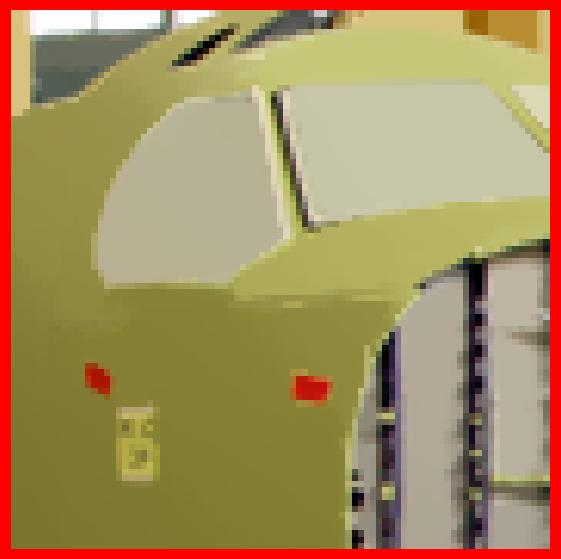}}\\
\subfloat[VDCNN]{\includegraphics[width=0.24\textwidth]{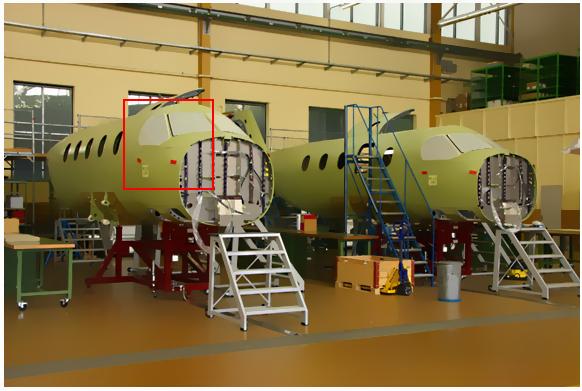}}
\hspace{-0.1\textwidth}\subfloat{\includegraphics[width=0.1\textwidth]{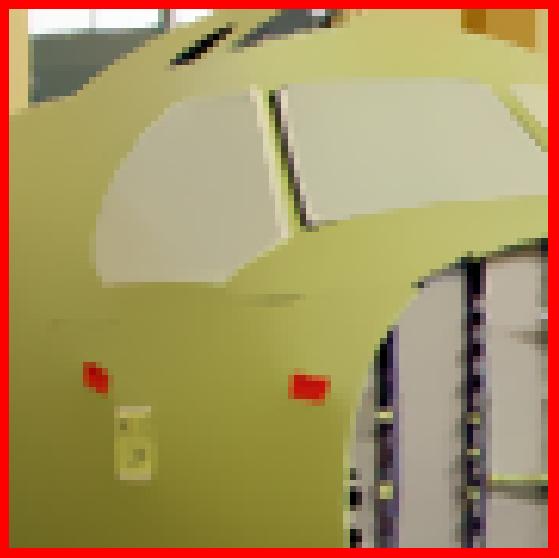}}
\hspace{0.001\textwidth}
\subfloat[ResNet]{\includegraphics[width=0.24\textwidth]{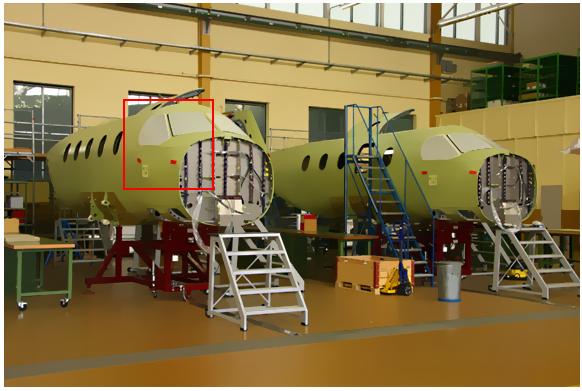}}
\hspace{-0.1\textwidth}\subfloat{\includegraphics[width=0.1\textwidth]{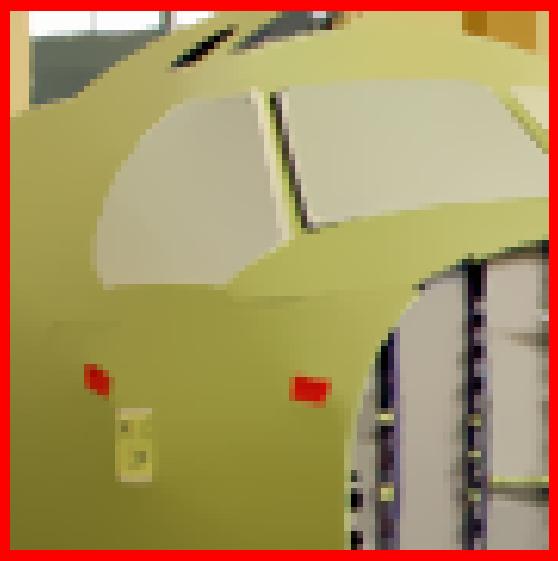}}
\caption{Comparison between $L_1$ smoothing algorithm and deep models.}
\label{fig:compare l1 ResNet}
\end{figure}

\subsection{Network Training}

We augment the training data with horizontal flips. RGB training patches are randomly sampled from source images and the corresponding smoothed images. We set different training patch sizes and mini-batch sizes for VDCNN and ResNet since they have different receptive fields and model complexity. For VDCNN, patch size is set to $41 \times 41$, and mini-batch size is set to 64. For ResNet, patch size is set to $96 \times 96$, and mini-batch size is set to 16.

Our deep models are trained using the ADAM optimizer \cite{kingma2014adam} with $\beta_1=0.9, \beta_2=0.999, \epsilon=10^{-8}$. The initial learning rate is set to $10^{-3}$. After the initial model converges, the learning rate is decreased by a factor of 10. Training is terminated once the model converges again.

We implemented the VDCNN and ResNet models in the Tensorflow framework and trained them using NVIDIA GeForce GTX 1080TI GPU. It takes one day to train VDCNN and two days to train ResNet, respectively.

\begin{figure*}[!t]
\captionsetup[subfigure]
  {subrefformat=simple, listofformat=subsimple,farskip=1pt}%labelformat=empty}
\centering
%\subfloat[Original image]{\includegraphics[width=0.23\textwidth]{origin_images//0422.png}}~
\subfloat{\includegraphics[width=0.23\textwidth]{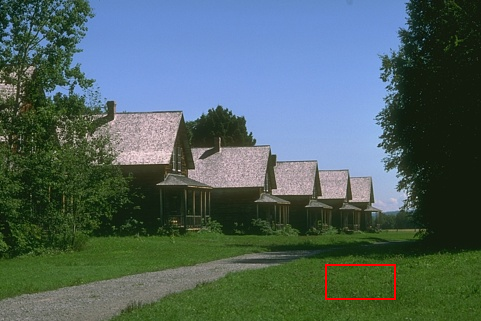}}~
\subfloat{\includegraphics[width=0.23\textwidth]{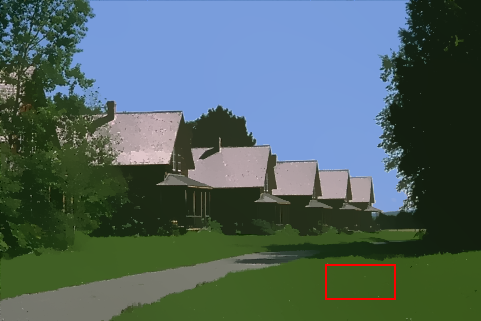}}~
\subfloat{\includegraphics[width=0.23\textwidth]{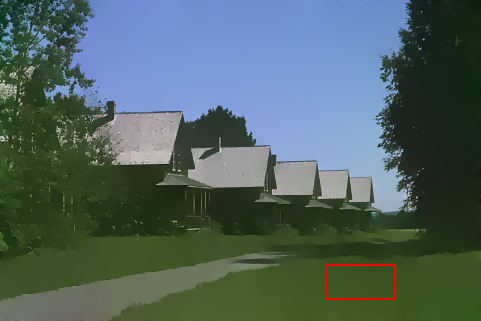}}~
\subfloat{\includegraphics[width=0.23\textwidth]{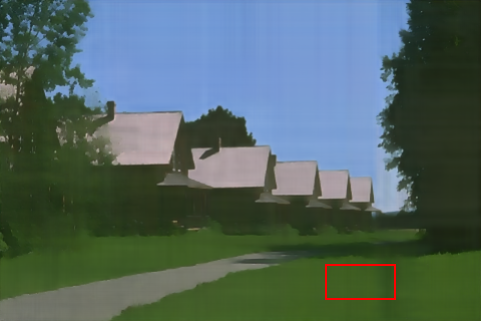}} \\
\addtocounter{subfigure}{-4}
\subfloat[Original image]{\includegraphics[width=0.23\textwidth]{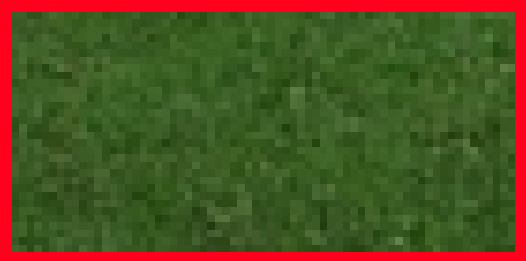}}~
\subfloat[Groundtruth image]{\includegraphics[width=0.23\textwidth]{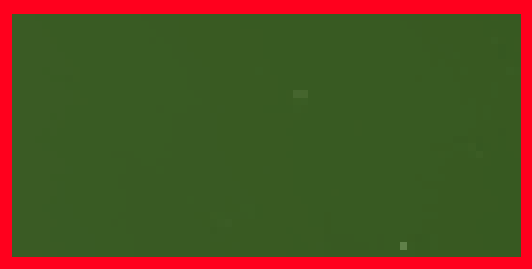}}~
\subfloat[Xu \textit{et al}.~\cite{xu2015deep}]{\includegraphics[width=0.23\textwidth]{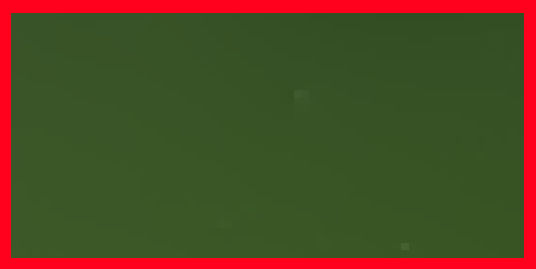}}~
\subfloat[Liu \textit{et al}.~\cite{liu2016learning}]{\includegraphics[width=0.23\textwidth]{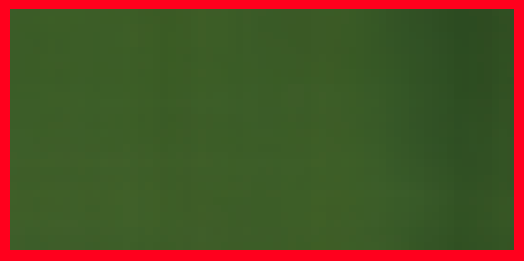}} \\
\addtocounter{subfigure}{-4}
\subfloat{\includegraphics[width=0.23\textwidth]{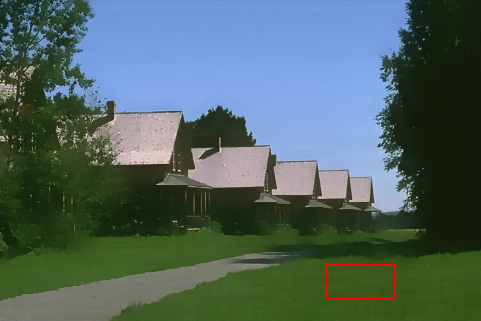}}~
\subfloat{\includegraphics[width=0.23\textwidth]{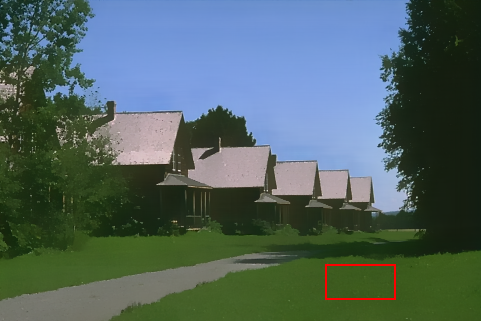}}~
\subfloat{\includegraphics[width=0.23\textwidth]{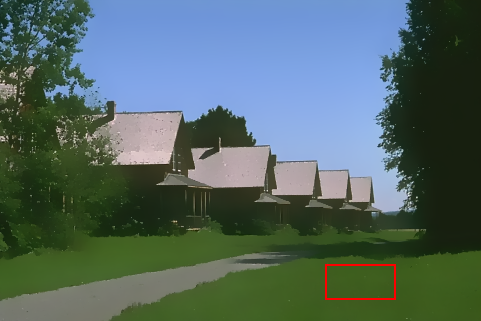}}~
\subfloat{\includegraphics[width=0.23\textwidth]{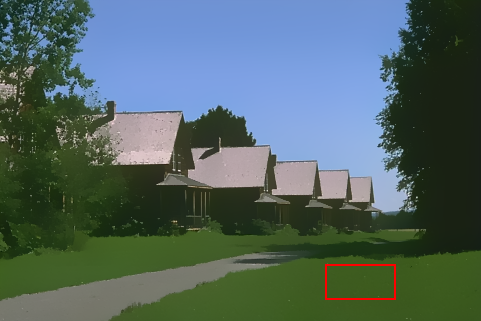}} \\
\subfloat[Li \textit{et al}.~\cite{li2016deep}]{\includegraphics[width=0.23\textwidth]{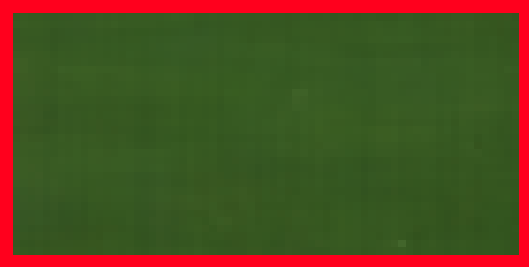}}~
\subfloat[Fan \textit{et al}.~\cite{fan2017generic}]{\includegraphics[width=0.23\textwidth]{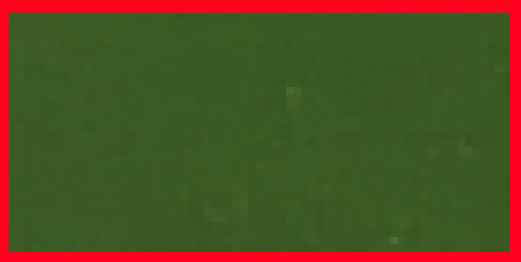}}~
\subfloat[VDCNN]{\includegraphics[width=0.23\textwidth]{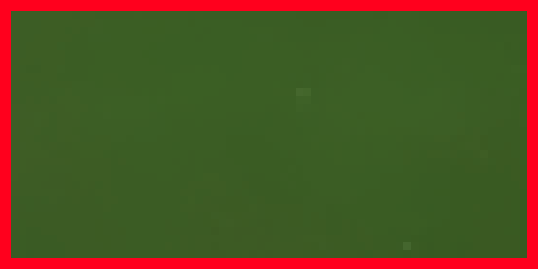}}~
\subfloat[ResNet]{\includegraphics[width=0.23\textwidth]{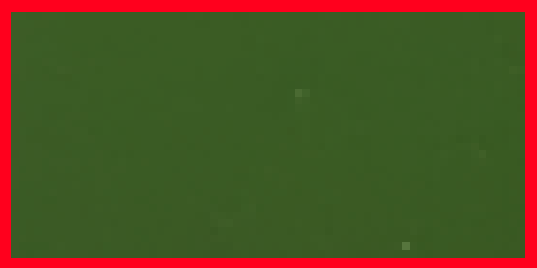}} \\
\caption{Qualitative comparison with recent CNN-based methods \cite{xu2015deep,liu2016learning,li2016deep,fan2017generic}. The groundtruth image shown in Figure (b) is the most frequently chosen image. It can be seen that \cite{xu2015deep} produces smoothed regions but suffers from color transitions. The method of \cite{liu2016learning} generates obvious artifacts at 'sky' and 'grass' regions. Methods in \cite{li2016deep} and \cite{fan2017generic} produce generally visually pleasing results but there still exist unwanted details. Our ResNet result is visually closer to the groundtruth. \textbf{Best viewed with zoom on screen.} }
\label{fig:CompareOtherCNN}
\end{figure*}

\subsection{Evaluation}

The VDCNN model and ResNet model trained with $loss_{l_1}+loss_{nb}$ are taken as the baseline algorithms for our edge-preserving smoothing benchmark. In this section, we report their performance both quantitatively and qualitatively.
%In this section, we demonstrate the effectiveness of the ResNet model trained using $loss_{l_1}+loss_{nb}$ both quantitatively and qualitatively.

As shown in Table~\ref{table:method evaluation}, the ResNet model achieves the lowest WRMSE and WMAE when compared to existing smoothing algorithms. The VDCNN model also achieves favorable performance, with slightly bigger errors than the ResNet model. An example is presented in Figure~\ref{fig:comparison model}, where our two baseline models are compared with the existing edge-preserving smoothing algorithms. Please note that these smoothing algorithms use their optimal parameter settings for achieving $\mbox{WMAE}^*$. It can be seen that some algorithms, such as SD filter, $L_0$ smoothing, Tree Filtering and LLF, cannot effectively remove trivial details at the floor regions. Some algorithms, such as FGS and WMF, blur the area around the foot regions with salient edges. $L_1$ smoothing and the deep models achieve overall better quality of edge-preserving smoothing than other algorithms.  More results can be found in the supplemental materials.

A more detailed visual comparison between the deep models and the $L_1$ smoothing algorithm \cite{bi20151} is given in Figure~\ref{fig:compare l1 ResNet} considering the fact that the $L_1$ smoothing algorithm is the most frequently chosen algorithm and it achieves the lowest WRMSE and WMAE among existing state-of-the-art algorithms. From Figure~\ref{fig:compare l1 ResNet} we can see that the $L_1$ smoothing algorithm wrongly increases the color contrast between two flattened regions on the airplane. The results from VDCNN and ResNet models do not have such artifacts.

As mentioned earlier, we do not aim to reproduce individual filters like \cite{xu2015deep,liu2016learning,li2016deep,fan2017generic}. By utilizing the constructed dataset, our baseline algorithm aims to train a deep CNN model that can produce reasonable edge-preserving smoothing results for a wide range of image contents without further tuning parameters. To the best of our knowledge, existing smoothing algorithms cannot perform consistently well on a wide range of image contents using a single parameter setting. As an example, a comparison between $L_0$ smoothing \cite{xu2011image} and our ResNet model is shown in Figure~\ref{fig:TuneParameter}. We can see that the $L_0$ smoothing algorithm needs to set different parameters for the `Racing car' and the `Gloves' images. If we set $\lambda$=0.03 for the `Racing Car' image, the edge between grass and road will blur. $\lambda$=0.01 is the proper setting for the `Racing Car' image. However, if we set $\lambda$=0.01 for the `Gloves' image, there still remain undesirable noises. In contrast, our ResNet model produces robust visual results on different images without tuning parameters. More results can be found in the supplementary file.

\begin{table}[!t]
\centering
\caption{Run time (second) of existing state-of-the-art edge-preserving smoothing algorithms and our deep models.}
\label{table:run time}
\resizebox{0.49\textwidth}{!}
{
\renewcommand{\arraystretch}{1.5}
\begin{tabular}{llllll}
\hline
\multicolumn{1}{|l|}{Method}   & \multicolumn{1}{l|}{SD filter}       & \multicolumn{1}{l|}{$L_0$ smoothing} & \multicolumn{1}{l|}{FGS}       & \multicolumn{1}{l|}{Tree Filtering} & \multicolumn{1}{l|}{WMF}  \\ \hline
\multicolumn{1}{|l|}{Run time} & \multicolumn{1}{l|}{10.46}           & \multicolumn{1}{l|}{1.24}            & \multicolumn{1}{l|}{0.05}      & \multicolumn{1}{l|}{0.18}           & \multicolumn{1}{l|}{0.52} \\ \hline
\multicolumn{1}{|l|}{Method}   & \multicolumn{1}{l|}{$L_1$ smoothing} & \multicolumn{1}{l|}{LLF}             & \multicolumn{1}{l|}{Our VDCNN} & \multicolumn{1}{l|}{Our ResNet}     & \multicolumn{1}{l|}{}     \\ \hline
\multicolumn{1}{|l|}{Run time} & \multicolumn{1}{l|}{328}             & \multicolumn{1}{l|}{199}             & \multicolumn{1}{l|}{0.41}      & \multicolumn{1}{l|}{0.78}           & \multicolumn{1}{l|}{}     \\ \hline
\end{tabular}
}
\end{table}

\subsection{Run Time}

In addition to visual quality, testing speed is also an important aspect for image smoothing methods. We report the running time of existing smoothing algorithms using the author-provided Matlab code on a 3.4GHz Intel i7 processor. The average running time over 100 testing images is shown in Table~\ref{table:run time}. The $L_1$ smoothing algorithm has lower $\mbox{WRMSE}^*$ and $\mbox{WMAE}^*$ than other smoothing algorithms, but spends hundreds of seconds on solving a series of large-scale sparse linear systems. The slow processing speed of $L_1$ smoothing algorithm prevents it from being an ideal pre-processing tool for other image processing applications, e.g., edge detection. In contrast, our ResNet-based model achieves the lowest WRMSE and WMAE while its GPU implementation runs faster than most existing state-of-the-art smoothing algorithms.

\subsection{Comparison with other CNN-based methods}

%Xu \textit{et al}.~\cite{xu2015deep}, Liu \textit{et al}.~\cite{liu2016learning}, Li \textit{et al}.~\cite{li2016deep}, Fan \textit{et al}.~\cite{fan2017generic} 

Recently, CNN-based approaches \cite{xu2015deep,liu2016learning,li2016deep,fan2017generic} have been proposed to reproduce individual smoothing filters. We also compare our method with those approaches on our proposed dataset. Note that \cite{xu2015deep,liu2016learning,li2016deep,fan2017generic} are originally designed to mimic smoothing filters where the groundtruth smoothed images are produced by the target filter. However, each source image in our dataset is associated with five groundtruth smoothed images and the quantitative measures are defined as weighted RMSE and weighted MAE (Equations \ref{eq:rmse2}-\ref{eq:mae2}). For fair comparison, we modify the loss function of previous CNN-based methods to weighted loss function like what we define in Equations \ref{eq:l2 loss}-\ref{eq:l1 loss}. 

\begin{table}[!t]
\centering
\caption{Quantitative comparison with recent CNN-based methods including Xu \textit{et al.}~\cite{xu2015deep}, Liu \textit{et al.}~\cite{liu2016learning}, Li \textit{et al}.~\cite{li2016deep} and Fan \textit{et al.}~\cite{fan2017generic}. Run time (second) of previous methods and our deep models are also reported.}
\label{table:compare_other_cnn}
\resizebox{0.49\textwidth}{!}
{
\renewcommand{\arraystretch}{1.5}
\begin{tabular}{|l|l|l|l|l|l|l|}
\hline
         & \cite{xu2015deep}   & \cite{liu2016learning}   & \cite{li2016deep}   & \cite{fan2017generic}  & Our VDCNN & Our ResNet \\ \hline
WRMSE    & 12.49 & 10.8  & 10.18 & 9.12  & 9.78      & 9.03       \\ \hline
WMAE     & 9.27  & 7.3   & 6.57  & 5.74  & 6.15      & 5.55       \\ \hline
Run time & 1.4  & 0.24 & 0.35 & 0.62 & 0.41     & 0.78     \\ \hline
\end{tabular}
}
\end{table}

Table \ref{table:compare_other_cnn} presents the quantitative results of our methods and previous CNN-based methods on the 100 test images. It can be seen that our ResNet-based model achieves the lowest $\mbox{WRMSE}$ and $\mbox{WMAE}$ thanks to the deeper structure and novel neighborhood loss function. There are totally 37 convolutional layers in our ResNet model. In comparison, Xu \textit{et al.}~\cite{xu2015deep} proposed a 3-layer convolutional neural networks to learn the gradient map. Liu \textit{et al.} incorporated convolutional neural networks in U-net style and recurrent neural networks together while the deep CNN consists of 9 layers. Li \textit{et al.}~\cite{li2016deep} proposed a joint network architecture of three components. Each component is a three-layer network. Fan \textit{et al}.~\cite{fan2017generic} presented a Cascaded Edge and Image Learning Network (CEILNet). Both E-CNN and I-CNN consist of 32 convolutional layers and residual unit is implemented for the middle layers.

Figure \ref{fig:CompareOtherCNN} shows some visual results of previous CNN-based approaches. It can be seen that \cite{xu2015deep} produces smoothed regions but it suffers from color transitions since this method requires a reconstruction step from gradient domain to final image output. The method of \cite{liu2016learning} generates obvious artifacts at 'sky' and 'grass' regions. Methods in \cite{li2016deep} and \cite{fan2017generic} produce generally visually pleasing results but there still exist unwanted details if we inspect closely. In contrast, our ResNet result is closer to the groundtruth.

\begin{figure*}[!t]
    \captionsetup[subfigure]
    {subrefformat=simple, listofformat=subsimple,farskip=3pt,labelformat=empty}
    \centering
    \subfloat{\includegraphics[height=0.20\textwidth]{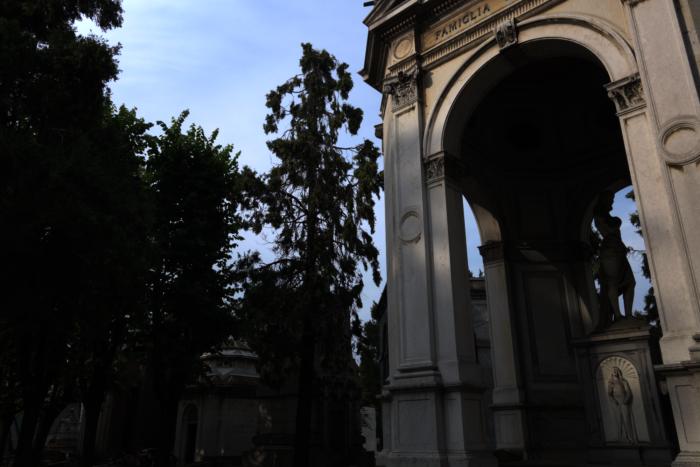}}~
    \subfloat{\includegraphics[height=0.20\textwidth]{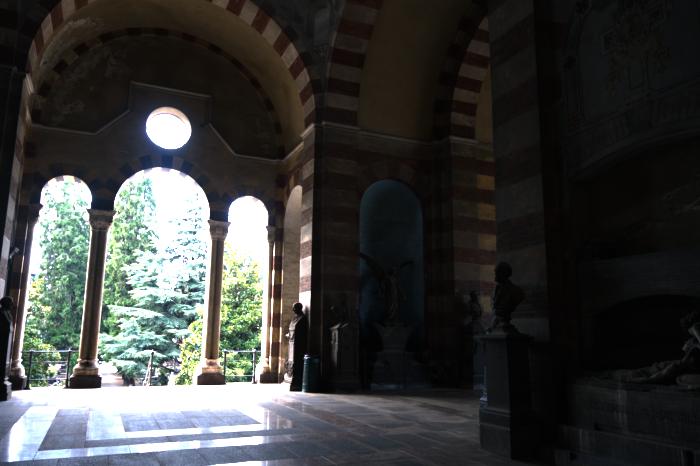}}~
    \subfloat{\includegraphics[height=0.20\textwidth]{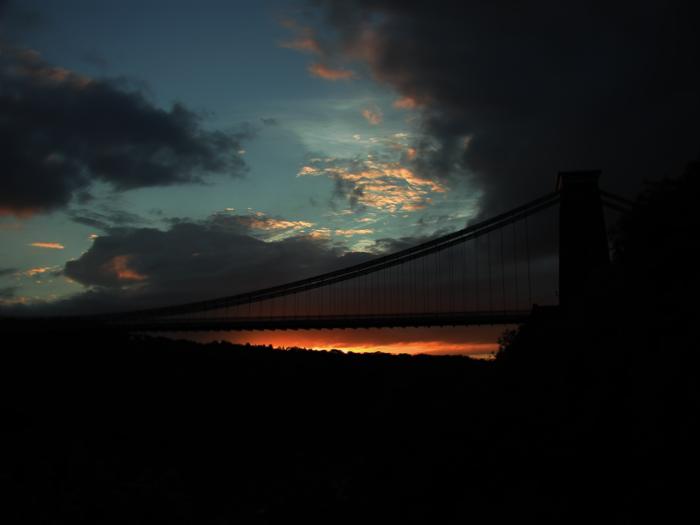}}\\
    \subfloat{\includegraphics[height=0.20\textwidth]{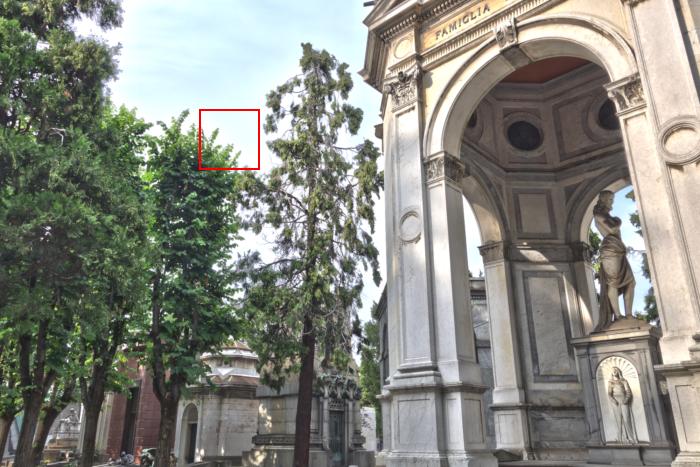}}
    \hspace{-0.1\textwidth}\subfloat{\includegraphics[height=0.1\textwidth]{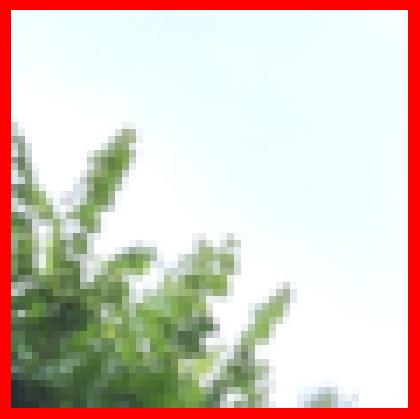}}~
    \subfloat{\includegraphics[height=0.20\textwidth]{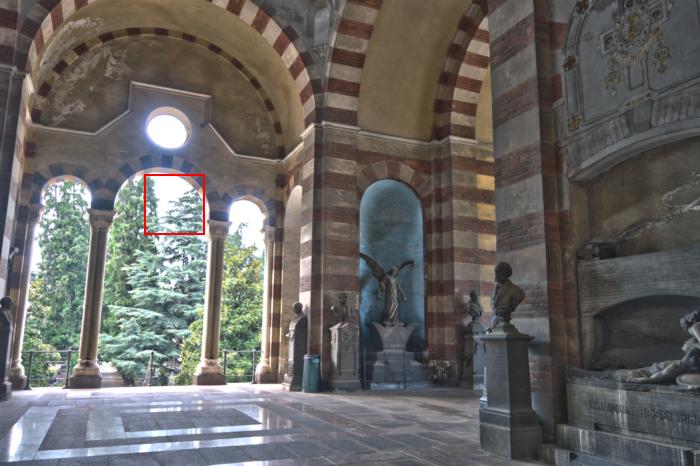}}
    \hspace{-0.1\textwidth}\subfloat{\includegraphics[height=0.1\textwidth]{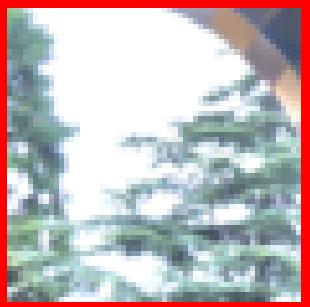}}~
    \subfloat{\includegraphics[height=0.20\textwidth]{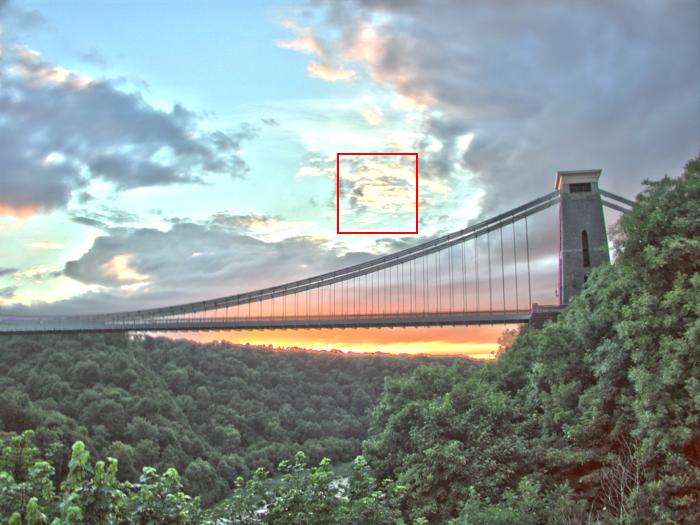}}
    \hspace{-0.1\textwidth}\subfloat{\includegraphics[height=0.1\textwidth]{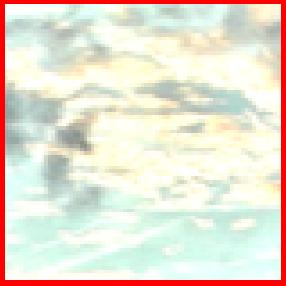}}\\
    \subfloat{\includegraphics[height=0.20\textwidth]{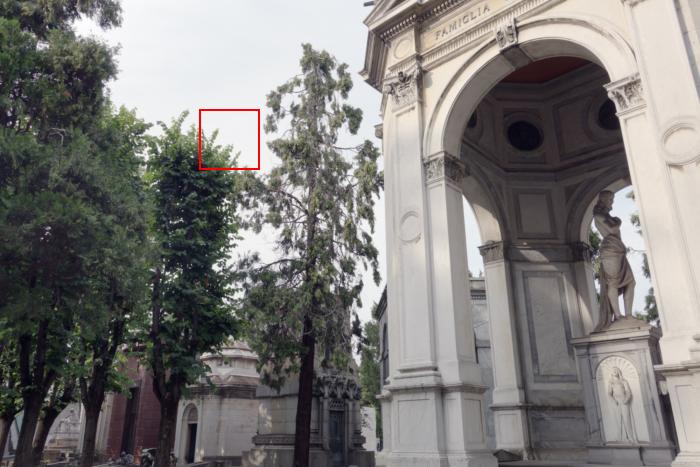}}
    \hspace{-0.1\textwidth}\subfloat{\includegraphics[height=0.1\textwidth]{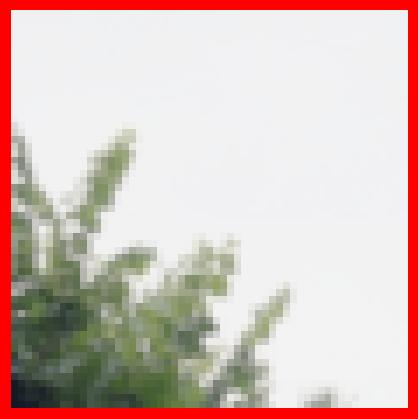}}~
    \subfloat{\includegraphics[height=0.20\textwidth]{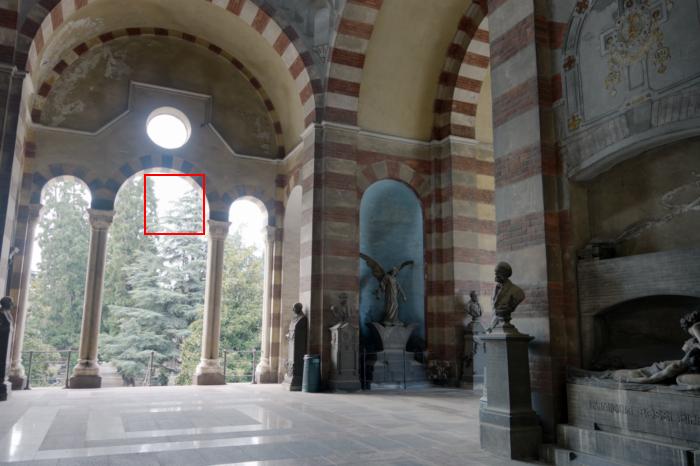}}
    \hspace{-0.1\textwidth}\subfloat{\includegraphics[height=0.1\textwidth]{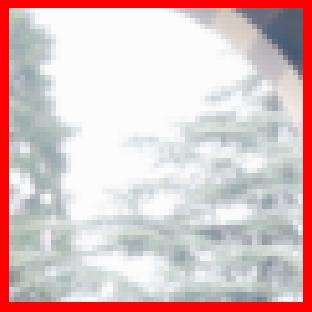}}~
    \subfloat{\includegraphics[height=0.20\textwidth]{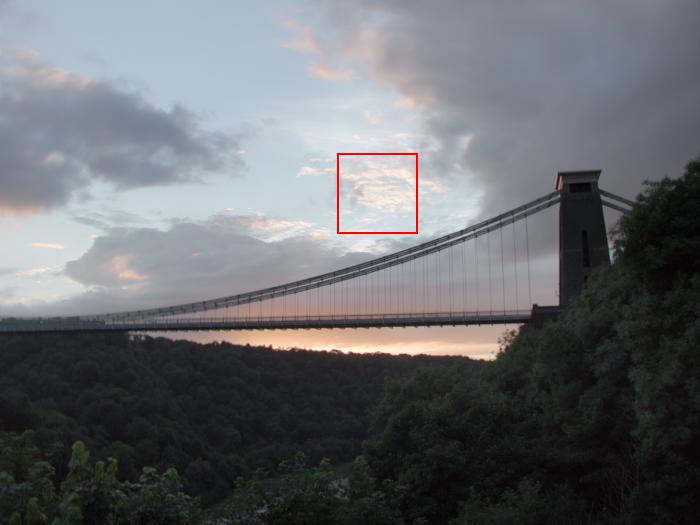}}
    \hspace{-0.1\textwidth}\subfloat{\includegraphics[height=0.1\textwidth]{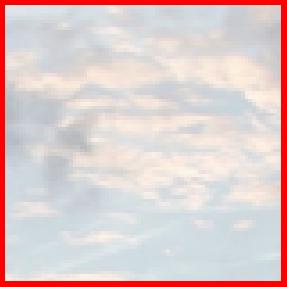}}\\
    \subfloat{\includegraphics[height=0.20\textwidth]{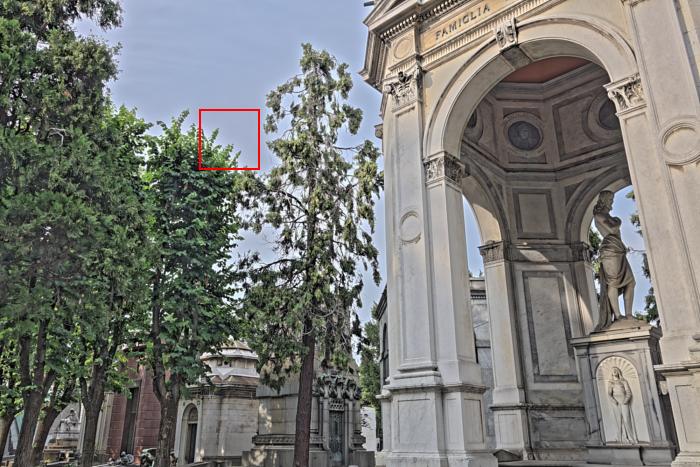}}
    \hspace{-0.1\textwidth}\subfloat{\includegraphics[height=0.1\textwidth]{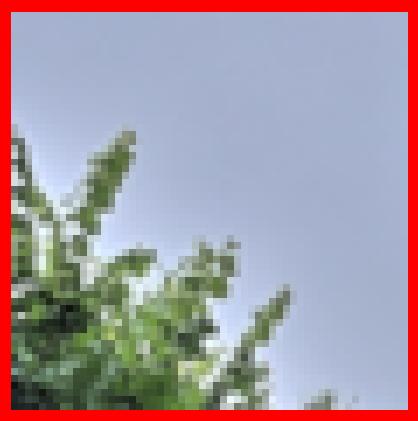}}~
    \subfloat{\includegraphics[height=0.20\textwidth]{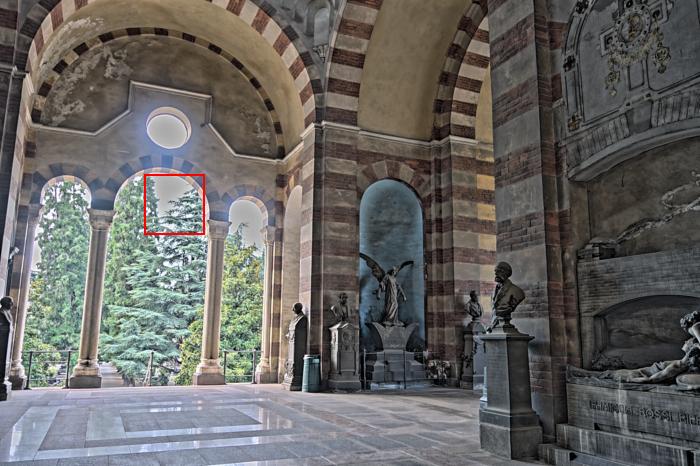}}
    \hspace{-0.1\textwidth}\subfloat{\includegraphics[height=0.1\textwidth]{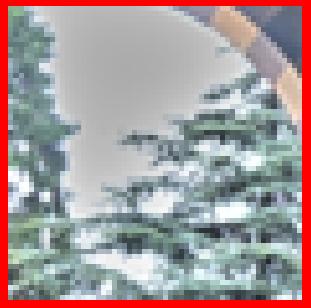}}~
    \subfloat{\includegraphics[height=0.20\textwidth]{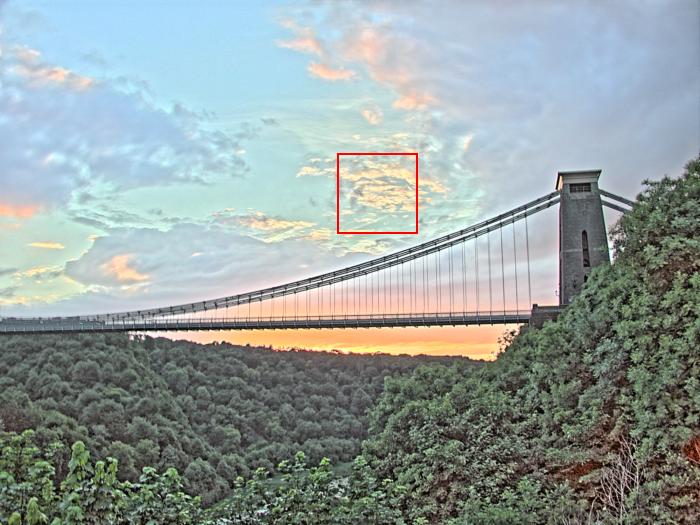}}
    \hspace{-0.1\textwidth}\subfloat{\includegraphics[height=0.1\textwidth]{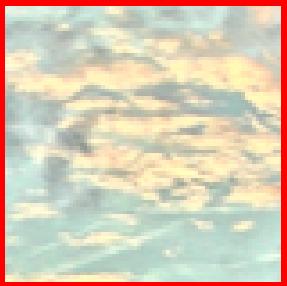}}\\
    \subfloat{\includegraphics[height=0.20\textwidth]{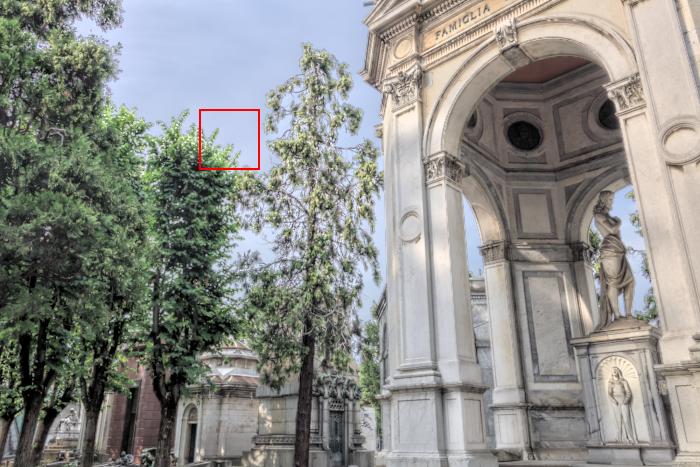}}
    \hspace{-0.1\textwidth}\subfloat{\includegraphics[height=0.1\textwidth]{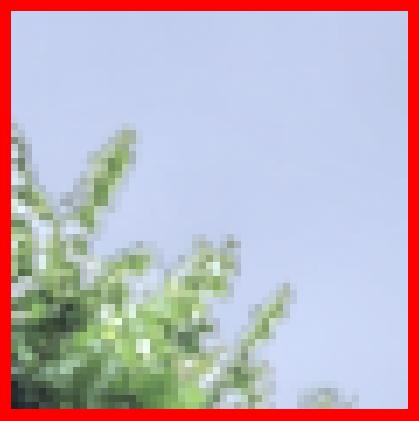}}~
    \subfloat{\includegraphics[height=0.20\textwidth]{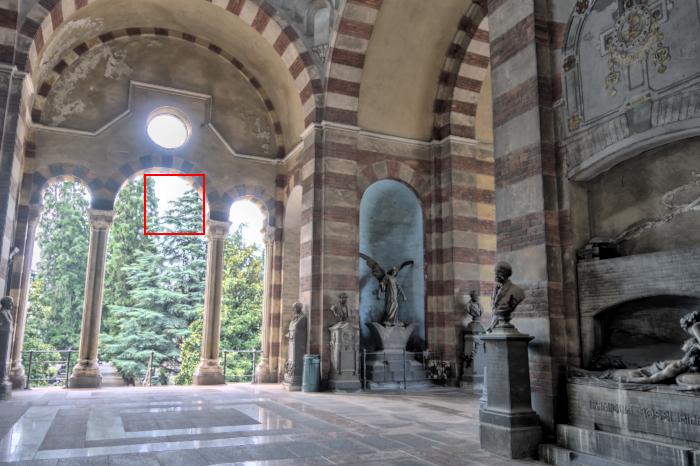}}
    \hspace{-0.1\textwidth}\subfloat{\includegraphics[height=0.1\textwidth]{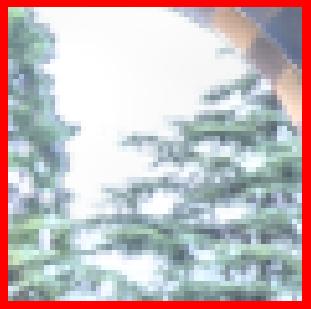}}~
    \subfloat{\includegraphics[height=0.20\textwidth]{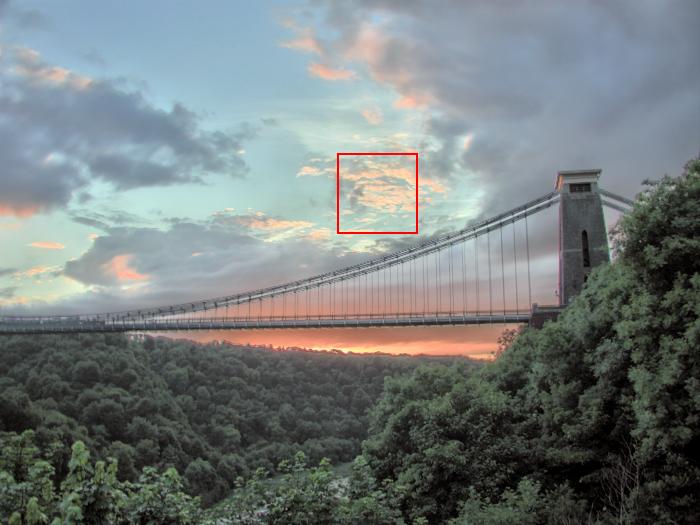}}
    \hspace{-0.1\textwidth}\subfloat{\includegraphics[height=0.1\textwidth]{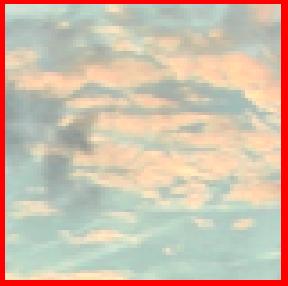}}
    \caption{Comparison of tone mapping results. \textbf{From top to bottom: the HDR images, results by BF method \cite{durand2002fast}, VAD method \cite{Ferradans2011}, LEP method \cite{Gu2013} and our ResNet model.} We can see that the BF method may introduce halo artifacts at the border areas around the tree in the top image. The results produced by VAD method miss many details. The LEP method over-enhances the images and leads to unnatural results. In contrast, our results preserve the details and look natural. An objective evaluation is shown in Table \ref{table:TMQI}.  }
    \label{fig:TMO}
\end{figure*}

\begin{figure*}[!t]
    \captionsetup[subfigure]
    {subrefformat=simple, listofformat=subsimple,farskip=2pt,labelformat=empty}
    \centering
    \subfloat{\includegraphics[height=0.2\textwidth]{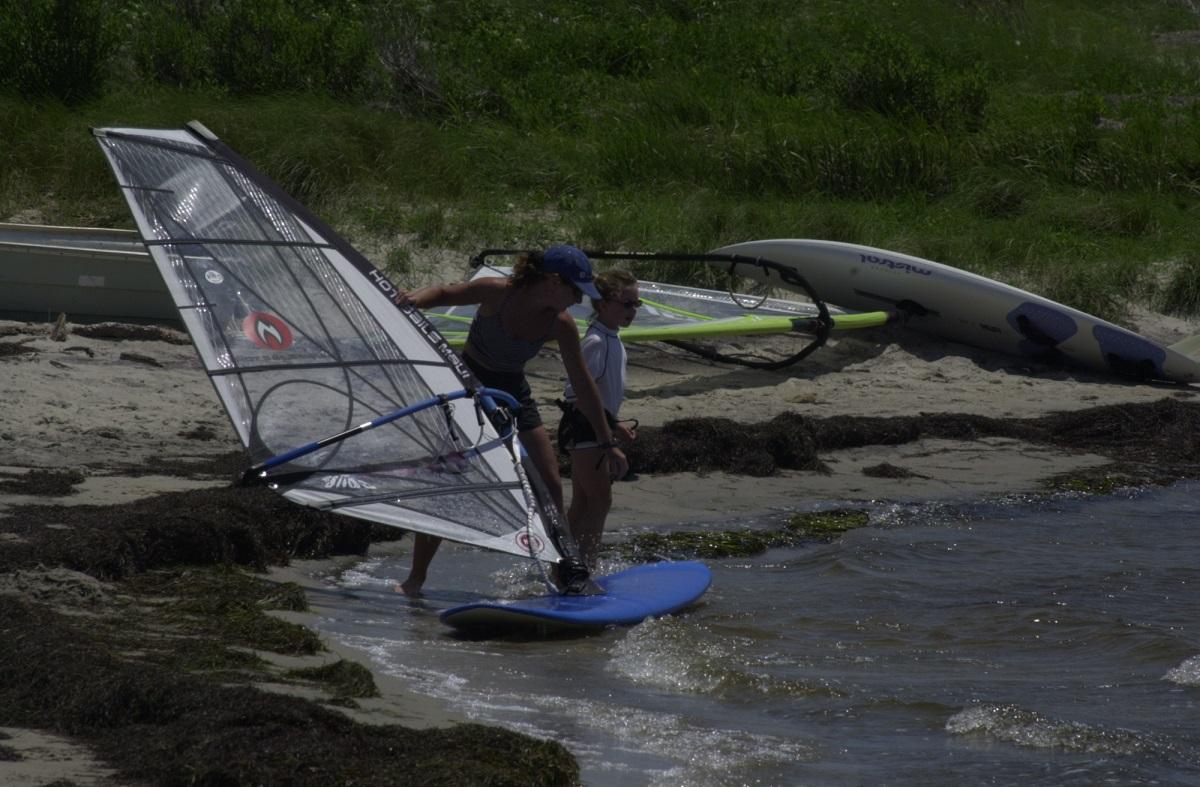}}~
    \subfloat{\includegraphics[height=0.2\textwidth]{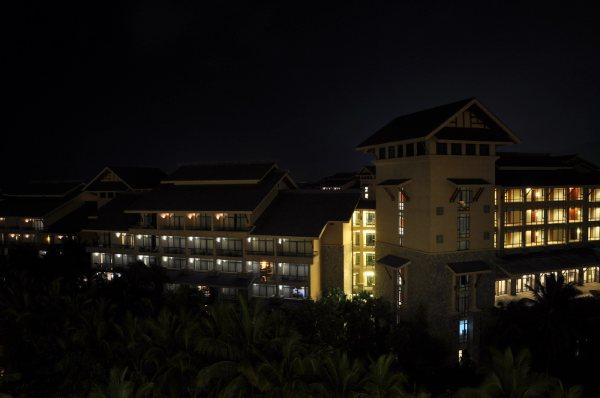}}~
    \subfloat{\includegraphics[height=0.2\textwidth]{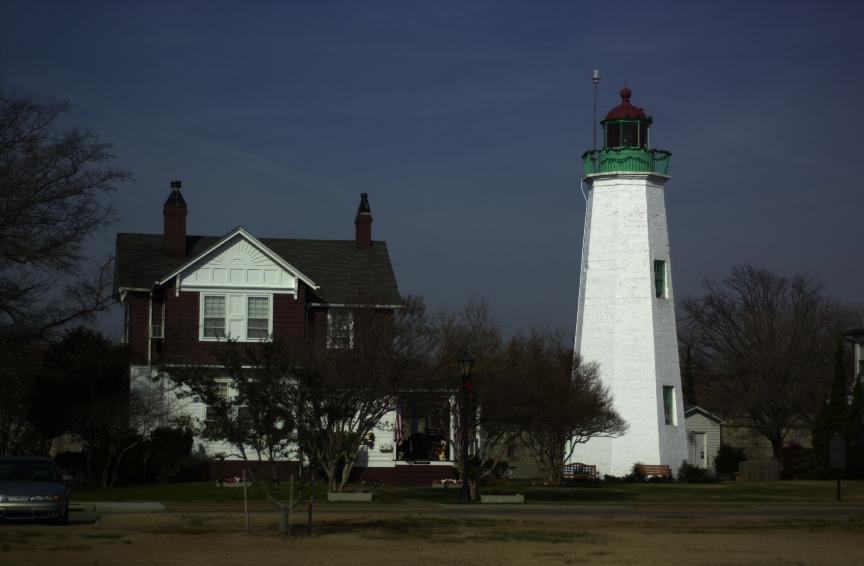}}\\
    \subfloat{\includegraphics[height=0.2\textwidth]{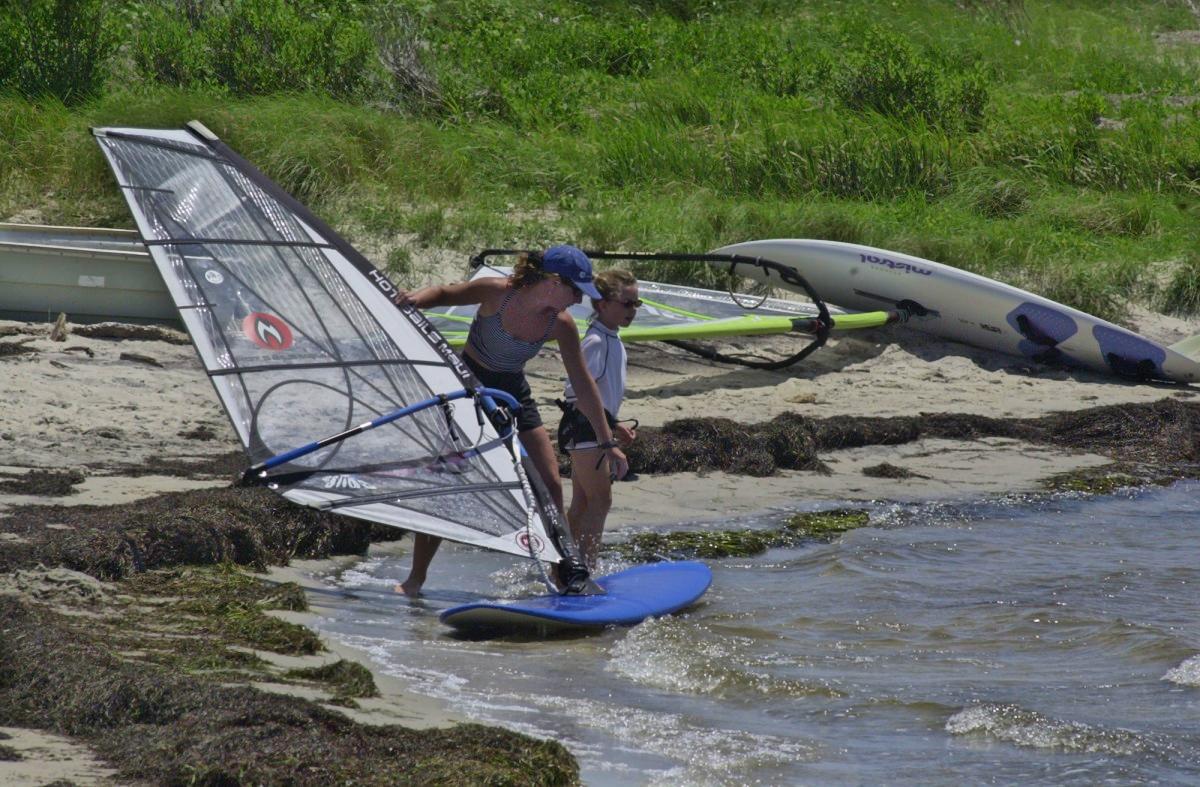}}~
    \subfloat{\includegraphics[height=0.2\textwidth]{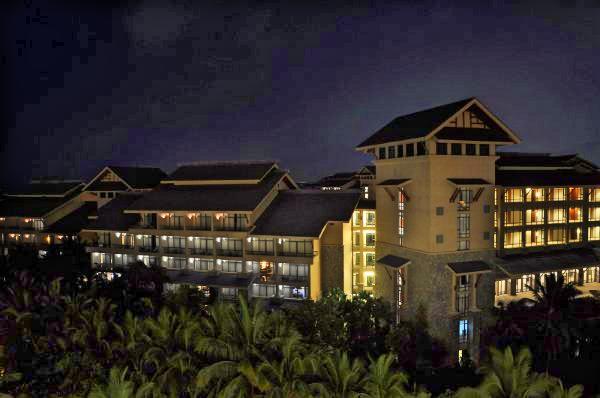}}~
    \subfloat{\includegraphics[height=0.2\textwidth]{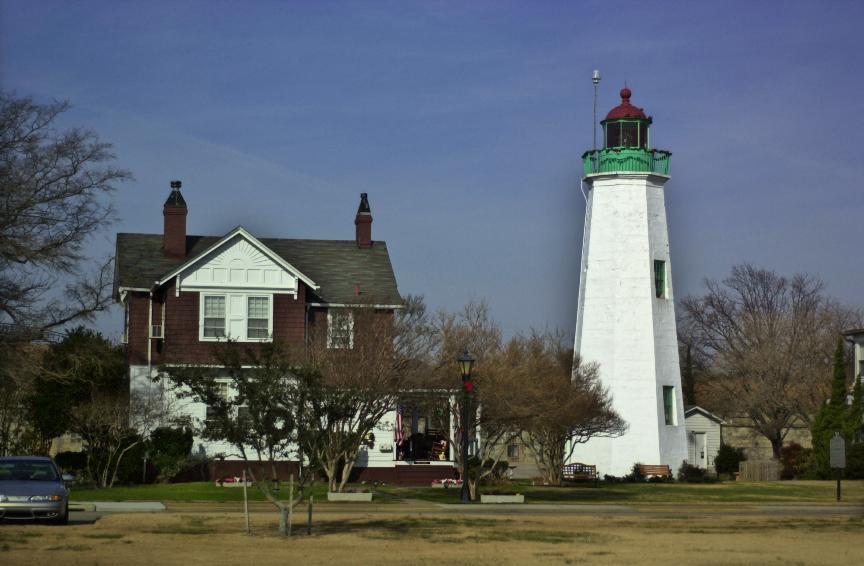}}\\
    \subfloat{\includegraphics[height=0.2\textwidth]{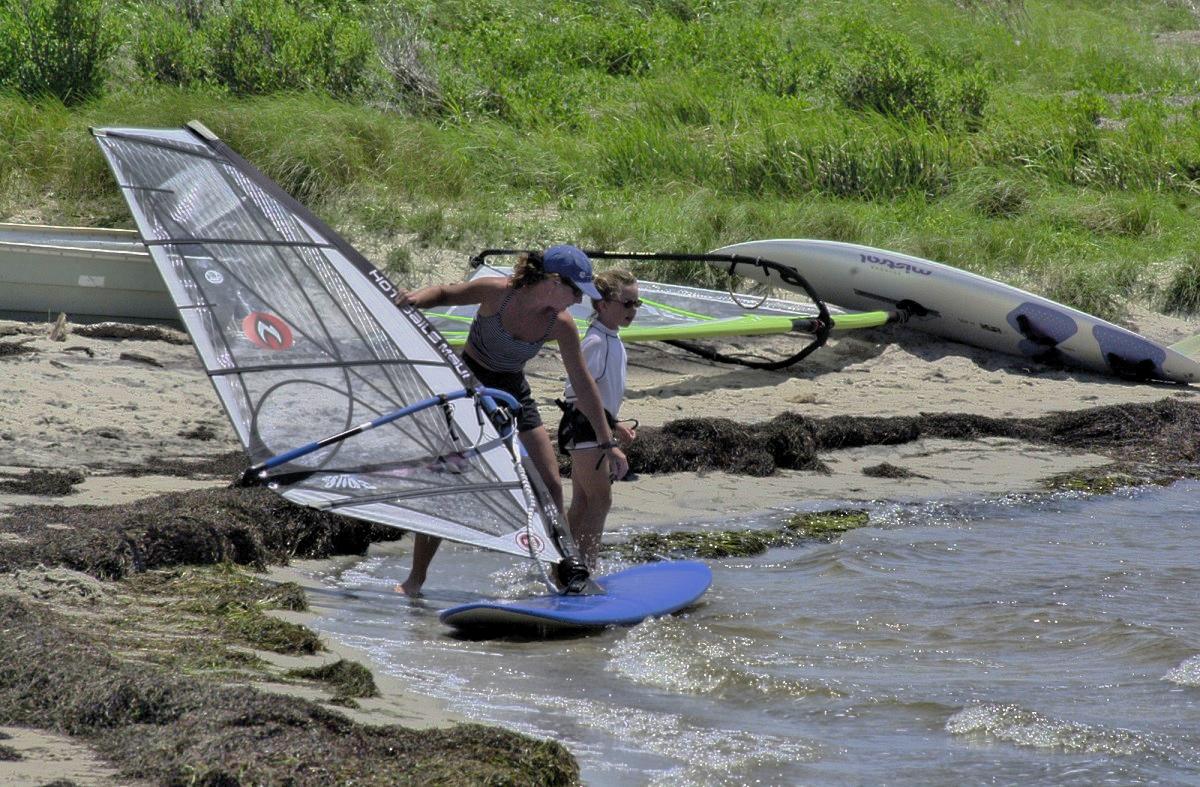}}~
    \subfloat{\includegraphics[height=0.2\textwidth]{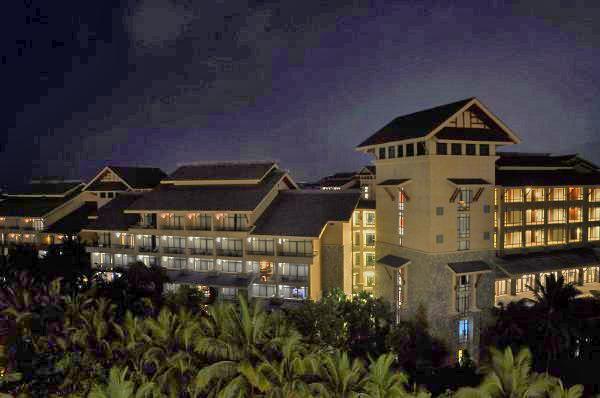}}~
    \subfloat{\includegraphics[height=0.2\textwidth]{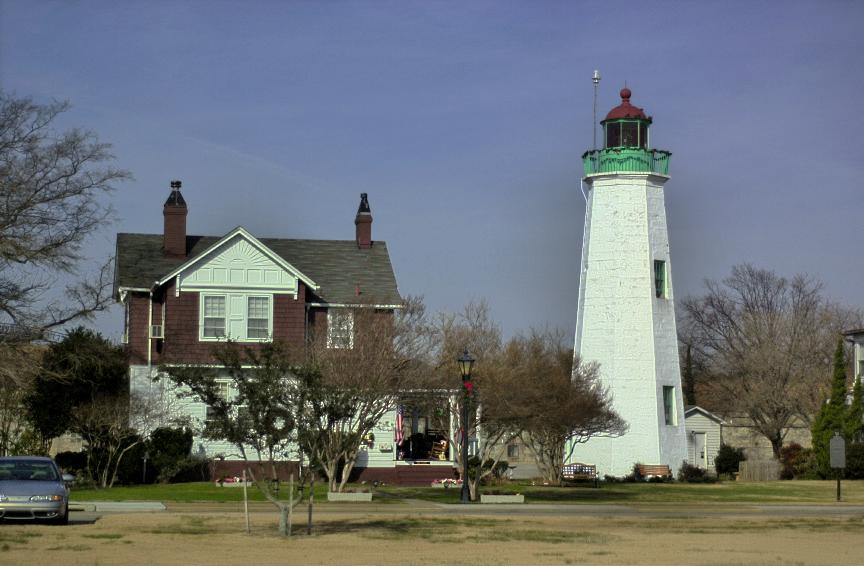}}\\
    \subfloat{\includegraphics[height=0.2\textwidth]{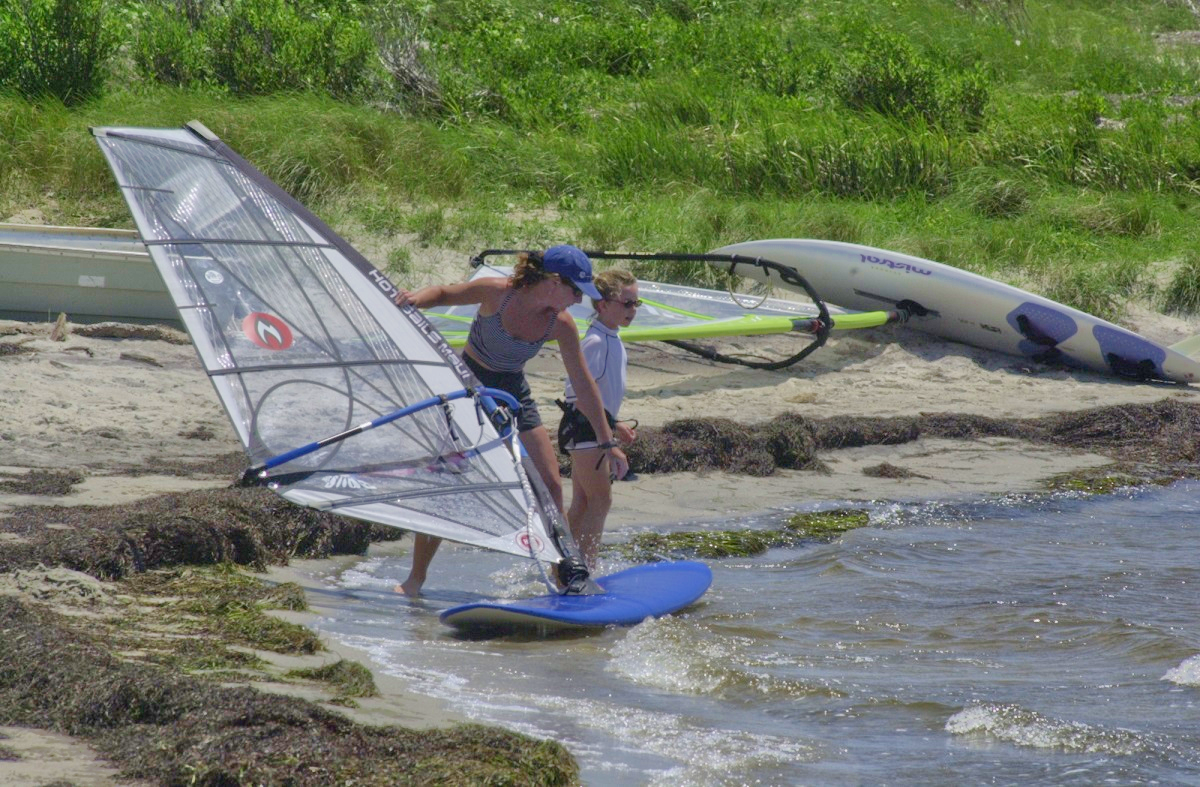}}~
    \subfloat{\includegraphics[height=0.2\textwidth]{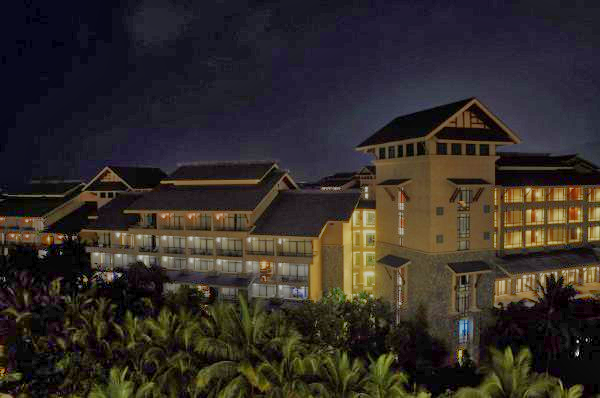}}~
    \subfloat{\includegraphics[height=0.2\textwidth]{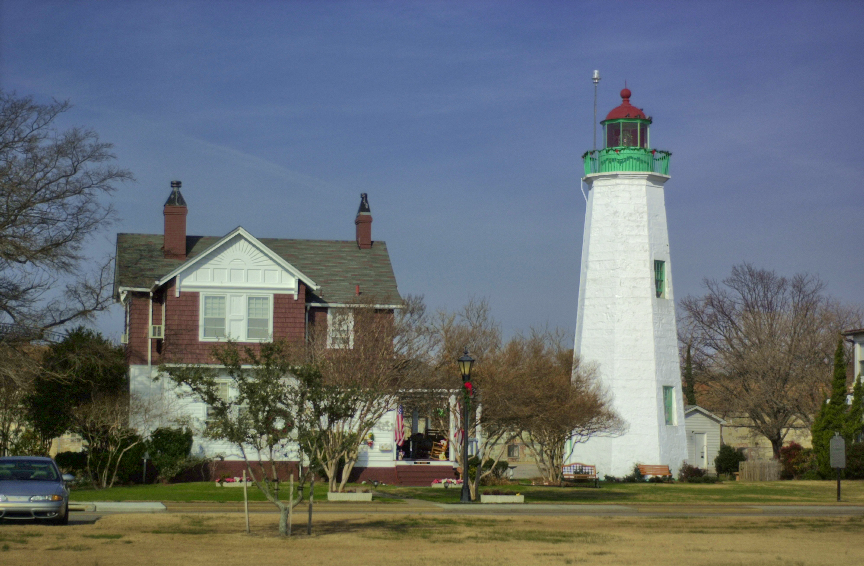}}\\  \subfloat{\includegraphics[height=0.2\textwidth]{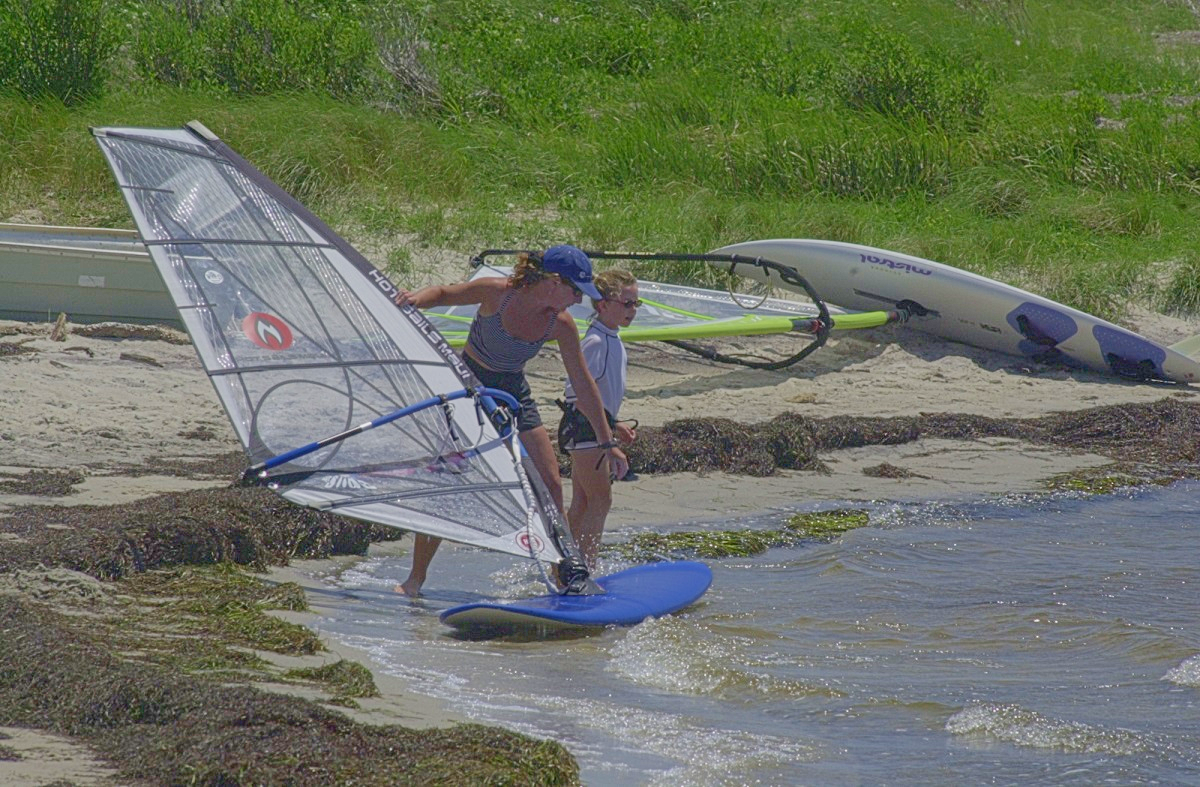}}~
    \subfloat{\includegraphics[height=0.2\textwidth]{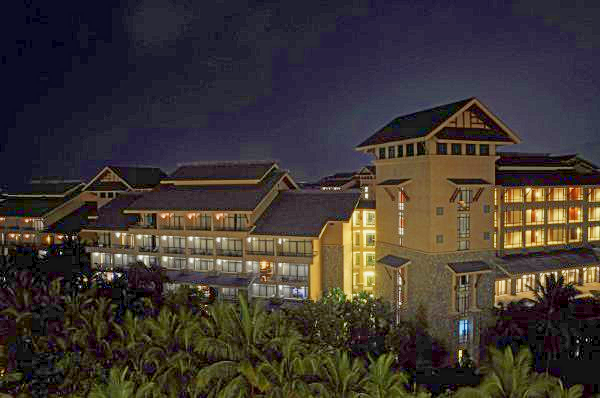}}~
    \subfloat{\includegraphics[height=0.2\textwidth]{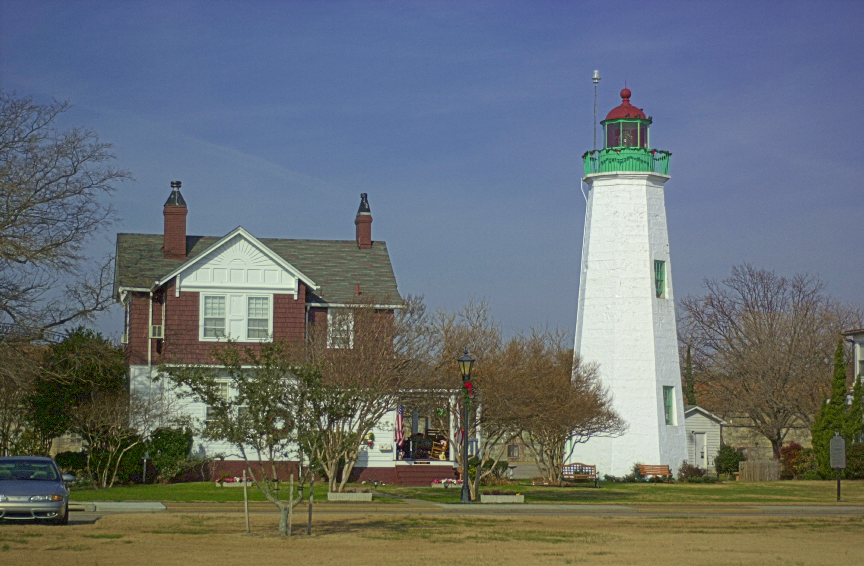}}\\
    \subfloat{\includegraphics[height=0.2\textwidth]{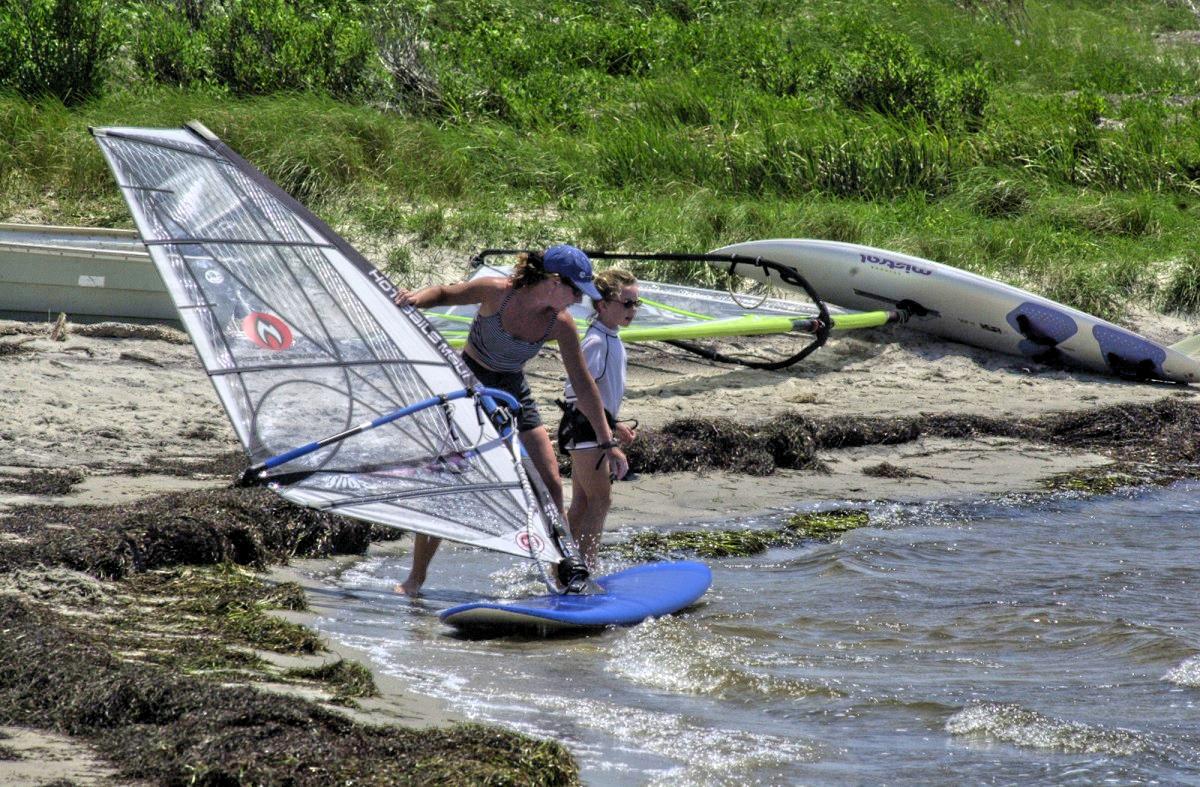}}~
    \subfloat{\includegraphics[height=0.2\textwidth]{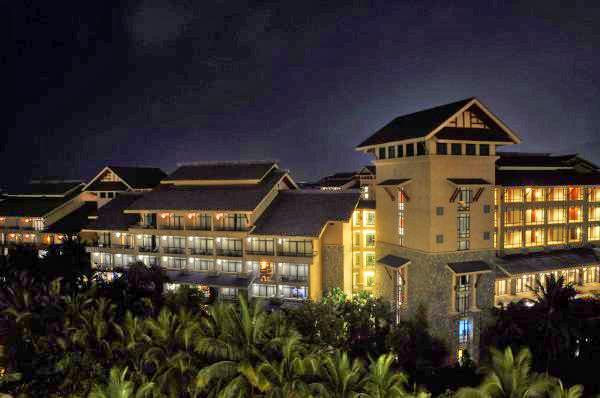}}~
    \subfloat{\includegraphics[height=0.2\textwidth]{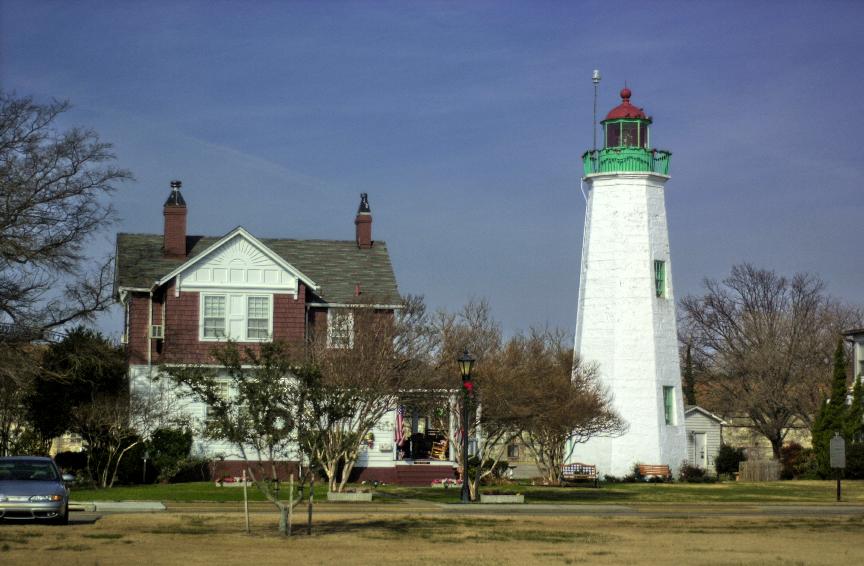}}
    \caption{Comparison of contrast enhancement results for low-light images. \textbf{From top to bottom: the original images, results by WV method \cite{Fu2016}, ND method \cite{Liang2016}, BF method \cite{tomasi1998bilateral}, GF method \cite{he2013guided} and our ResNet model.}  }
    \label{fig:CE}
\end{figure*}

\begin{table}[!t]
    \centering
    \caption{Comparison of TMQI scores. Tone Mapped image Quality Index (TMQI) \cite{Yeganeh2013} measures the structural fidelity and statistical naturalness.  }
    \label{table:TMQI}
    \resizebox{0.4\textwidth}{!}
    {
        \renewcommand{\arraystretch}{1.5}
        \begin{tabular}{|l|l|l|l|l|}
            \hline
            & BF\cite{durand2002fast} & VAD\cite{Ferradans2011} & LEP\cite{Gu2013} & Ours \\ \hline
            TMQI & 0.9020     & 0.9041            & 0.8750 & \textbf{0.9095}       \\   \hline
        \end{tabular}
    }
\end{table}

\begin{table}[!t]
\centering
\caption{Comparison of IEM. Image Enhancement Metric (IEM) \cite{jaya2013iem} measures the improvement in contrast of enhanced images.  }
\label{table:IEM}
\resizebox{0.45\textwidth}{!}
{
\renewcommand{\arraystretch}{1.2}
\begin{tabular}{|l|l|l|l|l|l|}
\hline
    & WV\cite{Fu2016}   & ND\cite{Liang2016}   & BF\cite{tomasi1998bilateral}   & GF\cite{he2013guided}   & Ours \\ \hline
IEM & 1.59 & 2.11 & 1.82 & 2.04 & \textbf{2.18} \\ \hline
\end{tabular}
}
\end{table}

\section{Applications}
\label{sec6}
Smoothing high-contrast details while preserving edges is a useful step in many applications \cite{durand2002fast,Ferradans2011,li2014visual,Gu2013,Liang2016,he2013guided}. In this section, we briefly discuss two applications, including tone mapping and contrast enhancement, by applying the trained ResNet model as the edge-preserving smoothing filter. 

\subsection{Tone mapping}

Tone mapping is a popular technique to map one set of colors to another to reproduce the appearance of a high dynamic range (HDR) image on a low dynamic range (LDR) displayer. The state-of-the-art tone mappers commonly adopt a layer decomposition scheme to decompose the HDR image into low- and high-frequency layers and then process them separately. In particular, the low frequency layer is estimated by applying an edge-preserving filter to the original HDR image. The edge-preserving property is very important for avoiding halo artifact and achieving naturalness in the tone-mapped images.Thus, a stable and effective edge-preserving filter is highly desirable to improve the tone mapping performance.

To avoid halo artifact, an edge-preserving filter should be able to preserve the strong edge regions and flatten other regions in the image, regardless of the image contents and types. Our ResNet baseline model can handle this task well, because it is trained on our dataset which is constructed with such criteria. We use the tone mapping framework in \cite{durand2002fast} by replacing the original bilateral filter by our ResNet model. We compare the tone mapped results with several state-of-the-art tone mappers, including bilateral filter method (BF) \cite{durand2002fast}, visual adaptation (VAD) \cite{Ferradans2011}, and local edge-preserving filter (LEP) \cite{Gu2013}. BF-based tone mapper \cite{durand2002fast} may not be as effective as the recently proposed approaches, but BF is widely adopted in different image processing tasks. On the other hand, VAD \cite{Ferradans2011} and LEP \cite{Gu2013} are selected because they obtain state-of-the-art performance. We do not compare with \cite{li2014visual} because saliency is beyond the scope of this work. Fig .\ref{fig:TMO} shows our tone mapping results compared with these tone mappers. We can see that our tone mapper with ResNet model reaches an excellent balance between halo removal and naturalness preservation. Other tone mappers suffers from either halo artifact or over-enhancement problems.

To better investigate the performance of the competitive tone mappers, we collected 100 HDR images online for objective evaluation. The TMQI metric \cite{Yeganeh2013} is used to score each tone mapped image by each method. Table \ref{table:TMQI} shows the average TMQI score of each tone mapper. We can see that our tone mapper with ResNet model achieves the highest TMQI score. This demonstrates that our edge-preserving benchmark can facilitate the tone mapper to gain robust performance over different image types and contents.

\subsection{Contrast enhancement}

Contrast enhancement aims to enhance the local contrast of an image that suffers from large illumination variation. Similar to tone mapping, a proper contrast enhancement framework decomposes an image into two components, illumination and reflectance. The estimation of illumination requires a high-performance edge-preserving filter. In \cite{Liang2016}, the authors proposed a criterion for optimal contrast enhancement, based on which the edge-preserving filter should preserve the boundary regions of an image and flatten the texture regions as much as possible. This coincides with the criterion adopted in our benchmark.

To test our benchmark in the application of contrast enhancement, we adopt the algorithm framework in \cite{Liang2016} and only replace the Diffusion-based filtering component by our ResNet model. We also compare with bilateral filter  \cite{tomasi1998bilateral} and guided filter (GF) \cite{he2013guided} since they have been widely used in different image processing tasks. The enhancement results together with the results of the state-of-the-art contrast enhancement methods including nonlinear diffusion method (ND) \cite{Liang2016} and weighted variational method (WV) \cite{Fu2016} are shown in Fig. \ref{fig:CE}. We can see that our enhancement results exhibit clearer structures and higher local contrast.

To measure the contrast enhancement results quantitatively, we report the Image Enhancement Metric (IEM) \cite{jaya2013iem}, a full reference IQA metric, to assess the contrast of the enhanced images. We used the 18 test images denoted as `a' to `r' in \cite{yue2017contrast} for evaluation. Some results are shown in Fig. \ref{fig:CE}, and all results can be found in supplementary. As Table \ref{table:IEM} shows, the improved performance validates that the ResNet model trained on our benchmark can be adopted as an efficient edge-preserving smoothing filter for image contrast enhancement.

\section{Conclusions}
We presented a benchmark for edge-preserving image smoothing for the purpose of quantitative performance evaluation and further advancing the state-of-the-art. This benchmark consists of 500 source images and their ``groundtruth" image smoothing results as well as baseline learning models. The baseline models are representative deep convolutional network architectures, on top of which we design novel loss functions well suited for edge-preserving image smoothing. Our trained deep networks run fast at test time while their smoothing results outperform state-of-the-art smoothing algorithms both quantitatively and qualitatively.

\section*{Acknowledgments}
This work was partially supported by Hong Kong Research Grants Council under General Research Funds (HKU17209714 and PolyU152124/15E). 

The authors would like to thank Lida Li, Jin Xiao, Xindong Zhang, Hui Li, Jianrui Cai, Sijia Cai, Hui Zeng, Hongyi Zheng, Wangmeng Xiang, Shuai Li, Runjie Tan, Nana Fan, Kai Zhang, Shuhang Gu, Jun Xu, Lingxiao Yang, Anwang Peng, Wuyuan Xie, Wei Zhang, Weifeng Ge, Kan Wu, Haofeng Li, Chaowei Fang, Bingchen Gong, Sibei Yang and Xiangru Lin for constructing the dataset, and the reviewers for their insightful comments.

\bibliographystyle{IEEEtran}
\normalem
\bibliography{egbib}

\begin{IEEEbiography}[{\includegraphics[width=1in,height=1.25in,clip,keepaspectratio]{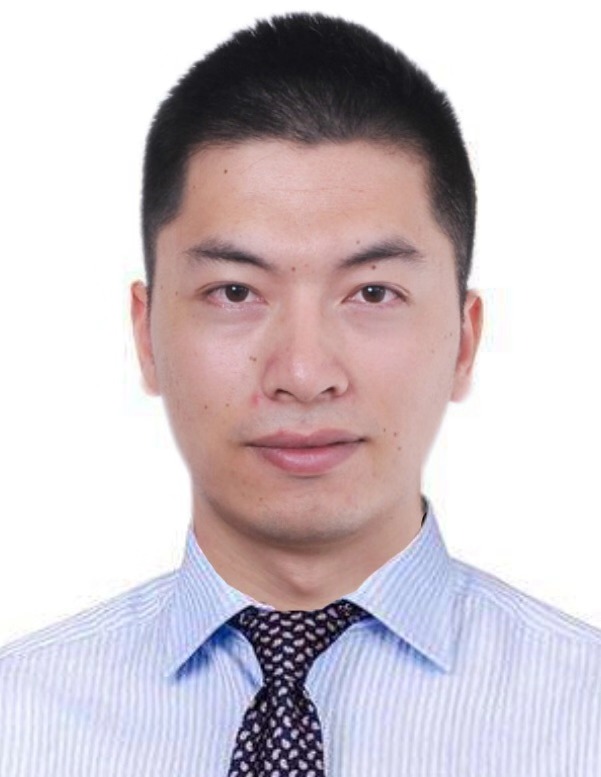}}]{Feida Zhu}
received the B.Eng. degree in the Department of Automation from The University of Science and Technology of China, Hefei, China in 2014, and Ph.D degree in the Department of Computer Science, The University of Hong Kong, Hong Kong, in 2019. He is now working as a research fellow in the School of Electrical and Electronic Engineering, Nanyang Technological University, Singapore. His research interests include image processing, computer vision and machine learning.
\end{IEEEbiography}

\begin{IEEEbiography}[{\includegraphics[width=1in,height=1.25in,clip,keepaspectratio]{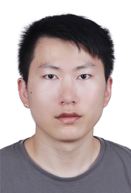}}]{Zhetong Liang}
received the B.S. degree in electronic information science and technology from the Guangdong University of Technology, Guangzhou, China, in 2013, and the M.S. degree with the School of Electronic and Information Engineering, South China University of Technology, Guangdong, China. He is currently pursuing the PhD degree with the Department of Computing, School of Engineering, the Hong Kong Polytechnic University. His current research interests include computational imaging, image processing pipeline and deep learning.
\end{IEEEbiography}

\begin{IEEEbiography}[{\includegraphics[width=1in,height=1.25in,clip,keepaspectratio]{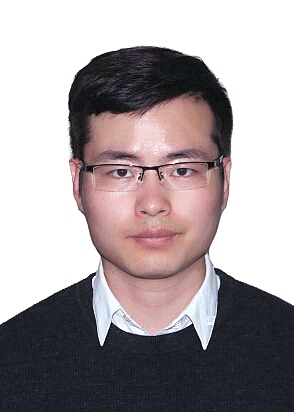}}]{Xixi Jia}
received the B.S., M.S. and PhD degrees from Xidian University, Xian, China, in 2012,2015 and 2018 respectively. He is now working at the School of Mathematics and Statistics, Xidian University. He was also worked as a research assistant at The Hong Kong Polytechnique University, HongKong from 2016-2017. His research interest covers image restoration, convex optimization and deep learning.
\end{IEEEbiography}

\begin{IEEEbiography}[{\includegraphics[width=1in,height=1.25in,clip,keepaspectratio]{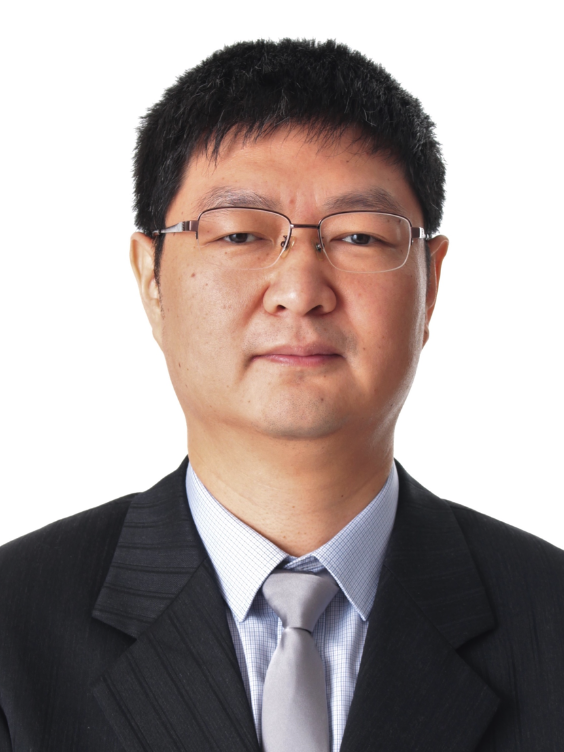}}]{Lei Zhang}
(M’04, SM’14, F’18) received his B.Sc. degree in 1995 from Shenyang Institute of Aeronautical Engineering, Shenyang, P.R. China, and M.Sc. and Ph.D degrees in Control Theory and Engineering from Northwestern Polytechnical University, Xi’an, P.R. China, in 1998 and 2001, respectively. From 2001 to 2002, he was a research associate in the Department of Computing, The Hong Kong Polytechnic University. From January 2003 to January 2006 he worked as a Postdoctoral Fellow in the Department of Electrical and Computer Engineering, McMaster University, Canada. In 2006, he joined the Department of Computing, The Hong Kong Polytechnic University, as an Assistant Professor. Since July 2017, he has been a Chair Professor in the same department. His research interests include Computer Vision, Image and Video Analysis, Pattern Recognition, and Biometrics, etc. Prof. Zhang has published more than 200 papers in those areas. As of 2019, his publications have been cited more than 39,000 times in literature. Prof. Zhang is a Senior Associate Editor of IEEE Trans. on Image Processing, and is/was an Associate Editor of SIAM Journal of Imaging Sciences, IEEE Trans. on CSVT, and Image and Vision Computing, etc. He is a “Clarivate Analytics Highly Cited Researcher” from 2015 to 2018. More information can be found in his homepage http://www4.comp.polyu.edu.hk/~cslzhang/.
\end{IEEEbiography}

\begin{IEEEbiography}[{\includegraphics[width=1in,height=1.25in,clip,keepaspectratio]{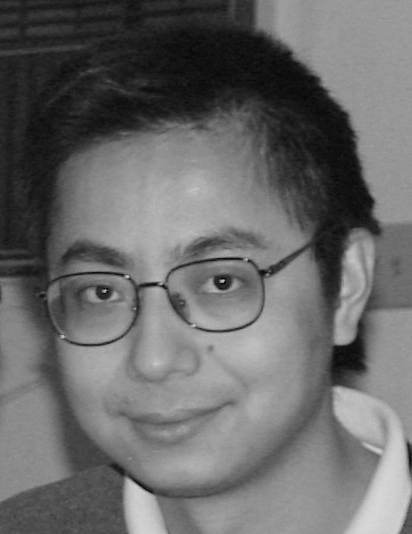}}]{Yizhou Yu}
(M'10, SM'12, F'19) received the PhD degree from University of California at Berkeley in 2000. He is a professor at The University of Hong Kong, and was a faculty member at University of Illinois at Urbana-Champaign for twelve years. He is a recipient of 2002 US National Science Foundation CAREER Award, 2007 NNSF China Overseas Distinguished Young Investigator Award, and ACCV 2018 Best Application Paper Award. Prof Yu has served on the editorial board of IET Computer Vision, The Visual Computer, and IEEE Transactions on Visualization and Computer Graphics. He has also served on the program committee of many leading international conferences, including SIGGRAPH, SIGGRAPH Asia, and International Conference on Computer Vision. His current research interests include computer vision, deep learning, biomedical data analysis, computational visual media and geometric computing.
\end{IEEEbiography}

\end{document}